\documentclass[journal]{IEEEtran}

\usepackage[utf8]{inputenc}
\usepackage{cite}
\usepackage{amsmath,amssymb,amsfonts}
\usepackage{algorithmic}
\usepackage{graphicx}
\usepackage{textcomp}
\usepackage{xcolor}
\usepackage{multirow}
\usepackage{booktabs}
\usepackage{colortbl}
\usepackage{tikz}
\usepackage{pgfplots}
\pgfplotsset{compat=1.17}
\usetikzlibrary{positioning}

\usepackage{CJKutf8}

\usepackage{pifont}   
\usepackage{amssymb}  
\newcommand{\cmark}{\checkmark}       
\newcommand{\pmark}{\ding{108}}       
\newcommand{\xmark}{\ding{55}}        

\definecolor{green}{RGB}{76,175,80}
\definecolor{orange}{RGB}{255,193,7}
\definecolor{red}{RGB}{244,67,54}

\begin{document}

\title{HiSciBench: A Hierarchical Multi-disciplinary Benchmark 
for Scientific Intelligence from Reading to Discovery}

\author{
Yaping Zhang\textsuperscript{1,2},
Qixuan Zhang\textsuperscript{1},
Xingquan Zhang\textsuperscript{1,2},
Zhiyuan Chen\textsuperscript{1,2},
Wenwen Zhuang\textsuperscript{1,2},
Yupu Liang\textsuperscript{1,2},\\
Lu Xiang,\textsuperscript{1,2}
Yang Zhao\textsuperscript{1,2,*},
Jiajun Zhang\textsuperscript{1,2},
Yu Zhou\textsuperscript{1,2},
and Chengqing Zong\textsuperscript{1,2}
\thanks{\textsuperscript{*}Corresponding author: Yang Zhao (email: yang.zhao@nlpr.ia.ac.cn).}
\thanks{\textsuperscript{1}Institute of Automation, Chinese Academy of Sciences.}
\thanks{\textsuperscript{2}University of the Chinese Academy of Sciences.}
}


\maketitle

\begin{abstract}
The rapid advancement of large language models (LLMs) and multimodal foundation models has sparked growing interest in their potential for scientific research. However, scientific intelligence encompasses a broad spectrum of abilities ranging from understanding fundamental knowledge to conducting creative discovery, and existing benchmarks remain fragmented. Most focus on narrow tasks and fail to reflect the hierarchical and multi-disciplinary nature of real scientific inquiry.  
We introduce \textbf{HiSciBench}, a hierarchical benchmark designed to evaluate foundation models across five levels that mirror the complete scientific workflow: \textit{Scientific Literacy} (L1), \textit{Literature Parsing} (L2), \textit{Literature-based Question Answering} (L3), \textit{Literature Review Generation} (L4), and \textit{Scientific Discovery} (L5). HiSciBench contains 8,735 carefully curated instances spanning six major scientific disciplines, including mathematics, physics, chemistry, biology, geography, and astronomy, and supports multimodal inputs including text, equations, figures, and tables, as well as cross-lingual evaluation.  
Unlike prior benchmarks that assess isolated abilities, HiSciBench provides an integrated, dependency-aware framework that enables detailed diagnosis of model capabilities across different stages of scientific reasoning. Comprehensive evaluations of leading models, including GPT-5, DeepSeek-R1, and several multimodal systems, reveal substantial performance gaps: while models achieve up to 69\% accuracy on basic literacy tasks, performance declines sharply to 25\% on discovery-level challenges.  
HiSciBench establishes a new standard for evaluating scientific Intelligence and offers actionable insights for developing models that are not only more capable but also more reliable. The benchmark will be publicly released to facilitate future research.
\end{abstract}

\begin{IEEEkeywords}
Scientific Intelligence,AI4Science,AI4Research
\end{IEEEkeywords}


\section{Introduction}

The rapid development of large language models (LLMs) has opened new possibilities for autonomous scientific research. While these models show impressive reasoning and language skills, evaluating their "scientific intelligence" remains a difficult and evolving challenge. Early benchmarks, such as MMLU~\cite{hendrycks2021measuring} and ScienceQA~\cite{lu2022scienceqa}, provided useful tests of basic scientific literacy. However, as LLMs become more advanced, these benchmarks have shown limitations in capturing the complex, multi-stage nature of real-world scientific work.

\begin{figure}[t]
\centering
\includegraphics[width=\linewidth]{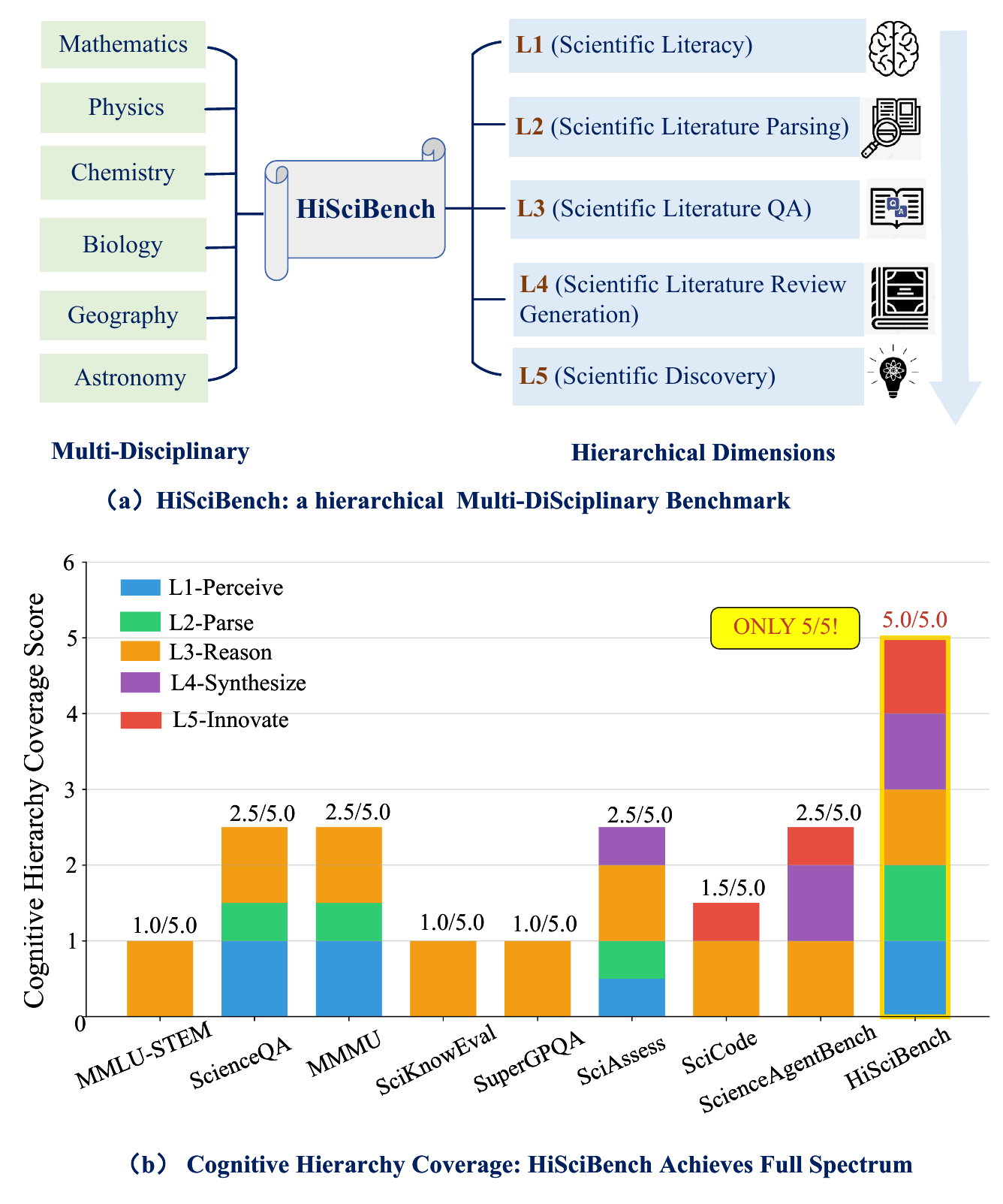}
\caption{
Overview of \textbf{HiSciBench}, a hierarchical benchmark for evaluating scientific intelligence in large language models.  
It covers six scientific disciplines including\textit{mathematics}, \textit{physics}, \textit{chemistry}, \textit{biology}, \textit{geography}, and \textit{astronomy}—shown on the left, 
and five progressively structured tasks (\textbf{L1–L5}) on the right, encompassing \textbf{Scientific Literacy}, \textbf{Literature Parsing}, \textbf{Literature QA}, \textbf{Review Generation}, and \textbf{Scientific Discovery}.  
The hierarchy mirrors the full scientific workflow, progressing from \textbf{L1} factual understanding and \textbf{L2} literature parsing, to \textbf{L3} contextual reasoning, \textbf{L4} integrative synthesis, and finally \textbf{L5} creative discovery.
}
\label{fig:framework}
\end{figure}

We identify two primary gaps in current evaluation methods.

\textbf{First, existing benchmarks are often fragmented and narrow in scope.} Most focus on isolated skills, such as factual recall~\cite{wang2024mmlupro}, visual reasoning~\cite{lu2024mathvista}, or specific problem-solving in chemistry or physics~\cite{wang2023scibench, mirza2024large}. Even comprehensive efforts like SciEval~\cite{lu2023scieval} tend to treat scientific ability as a flat set of independent tasks. In reality, scientific inquiry is hierarchical. According to Bloom’s Taxonomy~\cite{huber2025llms}, models often perform well on lower-order tasks like remembering facts but struggle with higher-order tasks like synthesizing information or creating new hypotheses. Currently, there is no unified benchmark that evaluates LLMs across the sequential stages of the actual research process.

\textbf{Second, existing evaluations lack sufficient multimodal and multilingual coverage.} Scientific research rarely relies on text alone; it integrates equations, figures, tables, and data across multiple languages. While some datasets have introduced multimodal elements~\cite{yue2024mmmu, lu2024mathvista}, they often focus on simple image recognition rather than the complex integration of cross-modal information found in scientific papers. Furthermore, most benchmarks are English-centric, failing to account for the global nature of modern science. As a result, they do not fully reflect the diverse inputs that researchers handle daily.

To address these limitations, we introduce \textbf{HiSciBench}, a Hierarchical multi-disciplinary Benchmark designed to evaluate scientific intelligence from reading to discovery. HiSciBench includes 8,735 curated instances across six core disciplines: mathematics, physics, chemistry, biology, geography, and astronomy.

Unlike previous "flat" evaluations, HiSciBench is organized into five levels (L1–L5) that follow the logical progression of a scientific workflow: \textbf{L1: Scientific Literacy} (factual knowledge and concepts).\textbf{L2: Literature Parsing} (multimodal document extraction and translation). \textbf{L3: Literature QA} (deep comprehension of specific papers).  \textbf{L4: Review Generation} (synthesizing information from multiple sources).  \textbf{L5: Scientific Discovery} (data-driven exploration and hypothesis generation). 
By structuring the benchmark this way, we can pinpoint exactly where models fail. Our testing shows that while state-of-the-art models perform well on foundational tasks, their performance drops significantly in the synthesis (L4) and discovery (L5) stages. This suggests that current AI is more capable of acting as a knowledge repository than as an active research assistant.

The main contributions of this work are as follows: \begin{itemize} \item We propose a hierarchical evaluation framework that moves beyond isolated tasks to reflect the actual multi-stage workflow of scientific research. \item We provide a multimodal and multilingual dataset of 8,735 instances across six disciplines, using authentic scientific literature to ensure high difficulty and realism. \item We conduct an extensive evaluation of leading LLMs, identifying key weaknesses in high-level reasoning and synthesis that provide a roadmap for future AI for Science development. \end{itemize}

\begin{table*}[!t]
\centering
\caption{Hierarchy of HiSciBench Cognitive Tasks and Examples. Each level reflects increasing cognitive complexity, from factual recall to scientific discovery.}
\renewcommand{\arraystretch}{1.2}
\setlength{\tabcolsep}{3pt}
\begin{tabular}{p{1.2cm} p{3.2cm} p{5.4cm} p{5.2cm}}
\hline
\textbf{Task ID} & \textbf{Task Name} & \textbf{Task Definition / Design Goal} & \textbf{Examples} \\
\hline
\rowcolor{green!10}
\multicolumn{4}{l}{\textbf{Level 1: Scientific Literacy.} \textit{Can the model RECALL and EXPLAIN fundamental scientific knowledge?}} \\

L1.1 & General Scientific Question Answering (QA)
& Evaluate factual recall and conceptual understanding across basic scientific disciplines such as physics, chemistry, and biology. 
& "What is the conservation of momentum?" \newline "Why does salt lower the freezing point of water?" \\

\hline
\rowcolor{green!10}
\multicolumn{4}{l}{\textbf{Level 2: Scientific Literature Parsing (OCR \& Translation).} \textit{Can the model READ and UNDERSTAND dense scientific text and visuals?}} \\

L2.1 & Literature OCR
& Recognize and extract text, tables, and formulas from scientific PDFs, images, figures, or charts, maintaining structural integrity. 
& "Read all textual and mathematical content from this figure and present the result in markdown format." \\

L2.2 & Literature Translation 
& Translate scientific documents across languages while preserving domain-specific terminology and semantic accuracy. 
& "Translate all the text in this image into Chinese and output in markdown format." \\

\hline
\rowcolor{green!10}
\multicolumn{4}{l}{\textbf{Level 3: Scientific Literature Question Answering (QA).} \textit{Can the model perform INFERENCE and REASONING within a single scientific document?}} \\

L3.1 & Monolingual Literature QA 
& Answer fine-grained questions within one scientific paper, integrating information from text, tables, and figures. 
& "What is the role of the 'Loss' component in the machine learning pipeline shown in the figure?" \newline "What method does the paper use to measure stability?" \\

L3.2 & Cross-lingual Literature QA 
& Perform question answering using a language different from that of the scientific document. 
& "\begin{CJK}{UTF8}{gbsn}问题4中讨论的图的直径是多少？\end{CJK} (What is the diameter of the graph discussed in Problem 4?)" \newline "\begin{CJK}{UTF8}{gbsn}当N=10时，$\gamma$的PRCC值是多少？\end{CJK} (What is the PRCC value of $\gamma$ when N=10?)" \\

\hline
\rowcolor{green!10}
\multicolumn{4}{l}{\textbf{Level 4: Scientific Literature Review Generation.} \textit{Can the model SYNTHESIZE knowledge across multiple papers?}} \\

L4.1 & Topic-guided Literature Review 
& Retrieve and summarize related works using provided keywords; organize findings into a coherent scientific overview. 
& "Generate a literature review on 'quantum computing with neutral atoms'." \newline "Summarize recent advances on machine learning for computational fluid dynamics." \\

\hline
\rowcolor{green!10}
\multicolumn{4}{l}{\textbf{Level 5: Scientific Discovery.} \textit{Can the model ANALYZE data and GENERATE novel hypotheses?}} \\

L5.1 & Data-driven Scientific Discovery 
& Given structured data and expert knowledge, analyze patterns, infer causal mechanisms, and propose plausible hypotheses. 
& "Based on the dataset of chemical reactions, predict new catalysts with higher yield." \newline "From experimental data, infer potential causal factors for material degradation." \\
\hline
\end{tabular}
\label{tab:HiSciBench_hierarchy}
\end{table*}

\begin{table*}[!t]
\centering
\caption{Comparison of HiSciBench with representative scientific AI benchmarks (2021--2025). HiSciBench provides comprehensive coverage of the 5 hierarchy (L1--L5), multilingual support, and innovation assessment. Size is reported as the total number of test instances.}
\label{tab:benchmark_comparison}

\resizebox{\textwidth}{!}{
\begin{tabular}{l|c|r|cc|ccccc|cc|c}
\toprule
\multirow{2}{*}{\textbf{Benchmark}} & \multirow{2}{*}{\textbf{Year}} & \multirow{2}{*}{\textbf{Size}} &
\multicolumn{2}{c|}{\textbf{Modality}} &
\multicolumn{5}{c|}{\textbf{Cognitive Hierarchy (Bloom's Taxonomy)}} &
\multicolumn{2}{c|}{\textbf{Advanced Features}} &
\multirow{2}{*}{\textbf{Disciplines}} \\
\cline{4-12}
& & & Text & Vision & L1-Perceive & L2-Parse & L3-Reason & \textbf{L4-Synthesize} & \textbf{L5-Innovate} & Multi-lingual & Agent/Code & \\ 
\midrule
\multicolumn{13}{l}{\textit{\textbf{General Scientific Understanding Benchmarks}}} \\
\midrule
MMLU-STEM~\cite{hendrycks2021mmlu} & 2021 & 3153 & \cmark & \xmark & \xmark & \xmark & \cmark & \xmark & \xmark & \xmark & \xmark & 4 \\
ScienceQA~\cite{lu2022scienceqa} & 2022 & 21208 & \cmark & \cmark & \cmark & \pmark & \cmark & \xmark & \xmark & \xmark & \xmark & 3 \\
SciBench~\cite{wang2023scibench} & 2023 & 692 & \cmark & \xmark & \xmark & \xmark & \cmark & \xmark & \xmark & \xmark & \xmark & 4 \\
GPQA~\cite{rein2023gpqa} & 2023 & 448 & \cmark & \xmark & \xmark & \xmark & \cmark & \xmark & \xmark & \xmark & \xmark & 3 \\
SciEval~\cite{lu2023scieval} & 2023 & 18045 & \cmark & \xmark & \xmark & \xmark & \cmark & \pmark & \xmark & \xmark & \xmark & 3 \\
MMMU~\cite{yue2024mmmu} & 2024 & 11500 & \cmark & \cmark & \cmark & \pmark & \cmark & \xmark & \xmark & \xmark & \xmark & 6 \\
SciKnowEval~\cite{feng2024sciknoweval} & 2024 & 28392 & \cmark & \xmark & \xmark & \xmark & \cmark & \xmark & \xmark & \xmark & \xmark & 4 \\
SuperGPQA~\cite{map2025supergpqa} & 2025 & 26529 & \cmark & \xmark & \xmark & \xmark & \cmark & \xmark & \xmark & \xmark & \xmark & Multi \\

\midrule
\multicolumn{13}{l}{\textit{\textbf{Domain-Specific Evaluation Benchmarks}}} \\
\midrule
MathVista~\cite{lu2024mathvista} & 2023 & 6141 & \cmark & \cmark & \cmark & \xmark & \cmark & \xmark & \xmark & \xmark & \xmark & 1 (Math) \\
ChemBench~\cite{mirza2024large} & 2024 & 4100 & \cmark & \pmark & \pmark & \xmark & \cmark & \xmark & \xmark & \xmark & \pmark & 1 (Chem) \\
BioASQ~\cite{DBLP:conf/aaaifs/TsatsaronisSPAAGGAAZN12} & 2015--23 & 4700+ & \cmark & \xmark & \xmark & \xmark & \cmark & \xmark & \xmark & \xmark & \xmark & 1 (Bio) \\
ChemLit-QA~\cite{wellawatte2025chemlit} & 2025 & 1054 & \cmark & \xmark & \xmark & \xmark & \cmark & \xmark & \xmark & \xmark & \xmark & 1 (Chem)\\
\midrule
\multicolumn{13}{l}{\textit{\textbf{Literature and Research-Oriented Benchmarks}}} \\
\midrule
SciRepEval~\cite{singh2023scirepeval} & 2023 & 6500 & \cmark & \xmark & \xmark & \pmark & \pmark & \pmark & \xmark & \xmark & \xmark & 5 \\
PaperQA~\cite{lala2023paperqa} & 2023 & 1000 & \cmark & \xmark & \xmark & \xmark & \cmark & \xmark & \xmark & \xmark & \xmark & Multi \\
SciAssess~\cite{cai2025sciassess} & 2024 & 10000 & \cmark & \pmark & \pmark & \pmark & \cmark & \pmark & \xmark & \xmark & \xmark & 4 \\
M3SciQA~\cite{li2024m3sciqa} & 2024 & 1400 & \cmark & \cmark & \cmark & \xmark & \cmark & \xmark & \xmark & \pmark & \xmark & 3 \\
ScholarBench~\cite{noh-etal-2025-scholarbench} & 2024 & 5200 & \cmark & \xmark & \xmark & \pmark & \cmark & \pmark & \xmark & \xmark & \xmark & Multi \\
AutoSurvey~\cite{wang2024autosurvey} & 2024 & 20 & \cmark & \xmark & \xmark & \xmark & \xmark & \cmark & \xmark & \xmark & \xmark & 1(CS) \\
ArXivQA~\cite{li2024multimodal} & 2024 & 100k & \cmark & \cmark & \xmark & \xmark & \cmark & \xmark & \xmark & \xmark & \xmark & Multi\\
Doc-750K~\cite{duan2025docopilot} & 2025 & 758k & \cmark & \cmark & \xmark & \pmark & \cmark & \xmark & \xmark & \xmark & \xmark & Multi \\
SFE~\cite{zhou2025scientists} & 2025 & 830 & \cmark & \cmark & \pmark & \xmark & \pmark & \xmark & \xmark & \xmark & \xmark & 5 \\
\midrule
\multicolumn{13}{l}{\textit{\textbf{Code Generation and Scientific Discovery}}} \\
\midrule
DS-1000~\cite{lai2023ds1000} & 2022 & 1000 & \cmark & \xmark & \xmark & \xmark & \cmark & \xmark & \xmark & \xmark & \cmark & 1 (DS) \\
AgentBench~\cite{liu2023agentbench} & 2023 & 8 envs & \cmark & \pmark & \pmark & \xmark & \cmark & \xmark & \xmark & \xmark & \cmark & Multi \\
SciCode~\cite{tian2024scicode} & 2024 & 338 & \cmark & \xmark & \xmark & \xmark & \cmark & \xmark & \pmark & \xmark & \cmark & 6 \\
SWE-bench~\cite{jimenez2024swebench} & 2024 & 2294 & \cmark & \xmark & \xmark & \xmark & \cmark & \xmark & \pmark & \xmark & \cmark & 1 (SE) \\
ScienceAgentBench~\cite{chen2025scienceagent} & 2025 & 102 & \cmark & \xmark & \xmark & \xmark & \cmark & \cmark & \pmark & \xmark & \cmark & 4 \\
\midrule
\rowcolor{green!10}
\textbf{HiSciBench (Ours)} & \textbf{2025} & \textbf{8735} & \textbf{\cmark} & \textbf{\cmark} & \textbf{\cmark} & \textbf{\cmark} & \textbf{\cmark} & \textbf{\cmark} & \textbf{\cmark} & \textbf{\cmark} & \textbf{\cmark} & \textbf{6} \\
\bottomrule
\end{tabular}
 }

\vspace{0.5em}

\begin{minipage}{\textwidth}
\footnotesize
\textit{Legend:} \cmark = fully supported; \pmark = partially supported; \xmark = not supported. \\
\textit{Cognitive Levels:} L1 (Perceive) = visual or textual perception and OCR. L2 (Parse) = document parsing and cross-lingual translation. L3 (Reason) = intra-document question answering and reasoning. L4 (Synthesize) = multi-document literature review generation. L5 (Innovate) = scientific discovery through code generation and problem solving. \\
\textit{Domains:} Math = Mathematics. Chem = Chemistry. Bio = Biology. DS = Data Science. SE = Software Engineering. Multi = multiple domains.
\end{minipage}

\end{table*}

\section{Related Work}

\subsection{General Scientific Understanding Benchmarks}

Early benchmarks like {MMLU}~\cite{hendrycks2021measuring} and its STEM subset evaluate scientific knowledge through multiple-choice questions but lack deeper reasoning capabilities. {ScienceQA}~\cite{lu2022scienceqa} introduces multimodal science questions with visual contexts, though limited to elementary-level reasoning. {SciBench}~\cite{wang2024scibench} addresses college-level problem-solving with 692 physics, chemistry, and mathematics problems, while {GPQA}~\cite{rein2023gpqa} provides 448 graduate-level questions requiring expert knowledge.

Recent large-scale efforts include {MMMU}~\cite{yue2024mmmu} with 11,500 multimodal college-level questions across six disciplines, {SciKnowEval}~\cite{feng2024sciknoweval} with 28,392 questions, and {SuperGPQA}~\cite{map2025supergpqa} with 26,529 expert-level questions. Domain-specific benchmarks such as {MathVista}~\cite{lu2024mathvista}, {ChemBench}~\cite{mirza2024large}, and {BioASQ}~\cite{DBLP:conf/aaaifs/TsatsaronisSPAAGGAAZN12} provide deep evaluation within particular fields. However, these benchmarks primarily focus on comprehension and reasoning (L3), overlooking higher-order cognitive processes such as synthesis (L4) and innovation (L5).

\subsection{Literature Understanding and Scientific Discovery}

Several benchmarks evaluate scientific literature understanding. {SciRepEval}~\cite{singh2023scirepeval} and {PaperQA}~\cite{lala2023paperqa} assess paper classification and question answering. {SciAssess}~\cite{cai2025sciassess} provides 10,000 instances across four domains with partial synthesis evaluation, while {M3SciQA}~\cite{li2024m3sciqa} introduces multi-document questions with visual reasoning. {ScholarBench}~\cite{noh-etal-2025-scholarbench} evaluates scholarly understanding with 5,200 instances, and {ArXivQA}~\cite{li2024multimodalarxiv} offers 100,000 multimodal questions from arXiv papers.

For code generation and scientific discovery, {DS-1000}~\cite{lai2023ds1000} evaluates data science coding, {SciCode}~\cite{tian2024scicode} targets scientific programming across six disciplines, and {ScienceAgentBench}~\cite{chen2025scienceagent} assesses autonomous agents with 102 tasks. While valuable, these benchmarks evaluate isolated capabilities rather than the complete research workflow, and lack systematic integration across cognitive levels.


Overall, existing benchmarks exhibit three fundamental limitations: (1) fragmented evaluation of isolated skills without complete workflow coverage, (2) limited multimodal and multilingual support, and (3) disconnection between agent-based evaluation and cognitive hierarchy. As shown in Table~\ref{tab:benchmark_comparison}, no benchmark fully supports all five cognitive levels from perception to innovation.


\label{subsec:dataset_stats}
\begin{figure*}[t]
    \centering
    \includegraphics[width=0.85\linewidth]{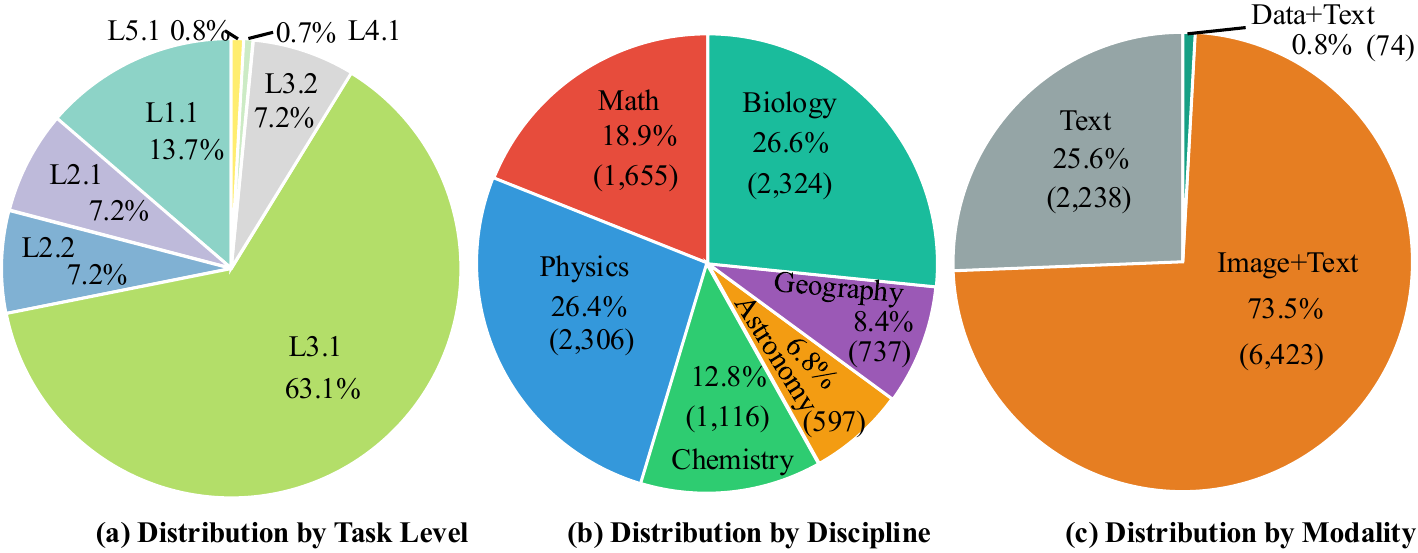}
    \caption{
    \textbf{HiSciBench: Comprehensive Distribution Analysis.} 
    (a) Distribution by task level across cognitive hierarchies (L1--L5), with perception and reasoning tasks (L3.1) dominating the benchmark. 
    (b) Distribution by scientific discipline, showing balanced coverage across six domains, where Biology (26.6\%) and Physics (26.4\%) are the largest. 
    (c) Distribution by modality, indicating that most tasks (84.7\%) involve structured image–text inputs, followed by text-only and data–text settings.
    }
    \label{fig:dataset_distribution}
\end{figure*}

\begin{table*}[!t]
\centering
\caption{Dataset statistics across HiSciBench levels. 
The benchmark spans six core scientific disciplines with diverse task formats ranging from factual QA to hypothesis generation.}
\label{tab:dataset_statistics}
\begin{tabular}{llp{5cm}cc}
\toprule
\textbf{Level} & \textbf{Task Name} & \textbf{Disciplines (Count)} & \textbf{Size} & \textbf{Data Source} \\
\midrule
L1.1 & General Scientific Question Answering (QA) & Math (200), Physics (200), Chemistry (200), Astronomy (200), Geography (200), Biology (200) & 1,200 & Public Datasets\textsuperscript{\dag} \\
\midrule
L2.1 & Literature OCR & Math (208), Physics (357), Astronomy (19), Biology (45) & 629 & Self-Collected\textsuperscript{\ddag} \\
\midrule
L2.2 & Literature Translation & Math (208), Physics (357), Astronomy (19), Biology (45) & 629 & Self-Collected\textsuperscript{\ddag} \\
\midrule
L3.1 & Monolingual Literature QA & Math (821), Physics (1,025), Chemistry (886), Astronomy (330), Geography (500), Biology (1,952) & 5,514 & Mixed\textsuperscript{\S} \\
\midrule
L3.2 & Cross-lingual Literature QA & Math (208), Physics (357), Astronomy (19), Biology (45) & 629 & Self-Collected\textsuperscript{\ddag} \\
\midrule
L4.1 & Topic-guided Literature Review & Math (10), Physics (10), Chemistry (10), Astronomy (10), Geography (10), Biology (10) & 60 & Self-Collected\textsuperscript{$\spadesuit$} \\
\midrule
L5.1 & Data-driven Scientific Discovery & Chemistry (20), Geography (27), Biology (27) & 74 & Public Dataset\textsuperscript{$\P$} \\
\midrule
\textbf{Total} & \textbf{7 Tasks} & \textbf{6 Disciplines} & \textbf{8,735} & \\
\bottomrule
\end{tabular}

\vspace{0.3em}
\footnotesize
\textsuperscript{\dag}Sampled from SuperGPQA benchmarks. \\
\textsuperscript{\ddag}Collected from arXiv LaTeX sources and rendered as PDF images. \\
\textsuperscript{\S}Aggregated from ScholarChemQA, arXivQA, PubMedQA, and augmented with self-annotated examples. \\
\textsuperscript{$\spadesuit$}Collected from arXiv and Nature publications. \\
\textsuperscript{$\P$}Adapted from ScienceAgentBench benchmark.
\end{table*}

\section{HiSciBench Benchmark Design}
\label{sec:benchmark_design}
In this section, we present the comprehensive design of HiSciBench, a hierarchical benchmark system that systematically evaluates multimodal large language models across the full spectrum of scientific research capabilities. Our design philosophy centers on simulating the cognitive progression inherent in human scientific inquiry, spanning from foundational knowledge acquisition to advanced research innovation.

\subsection{Dataset Statistics and Characteristics}
Figure~\ref{fig:dataset_distribution} and Table~\ref{tab:dataset_statistics} provide a comprehensive overview of the HiSciBench dataset. 
HiSciBench spans {five different levels (L1--L5)} across {six scientific disciplines}, covering diverse modalities such as text, image–text, and structured data. 
This hierarchical design enables systematic evaluation of models' scientific reasoning—from factual recall to data-driven discovery.

As shown in Figure~\ref{fig:dataset_distribution}(a), Level~3 tasks (\textit{Monolingual and Cross-lingual Literature QA}) dominate the dataset, accounting for 63\% of all instances. 
This reflects HiSciBench’s focus on reasoning within real-world scientific literature. 
In terms of disciplinary coverage (Figure~\ref{fig:dataset_distribution}b), {Biology (26.6\%)} and {Physics (26.4\%)} constitute the largest portions, followed by Mathematics and Chemistry, ensuring balanced representation across major scientific domains. 
The modality analysis in Figure~\ref{fig:dataset_distribution}(c) shows that most tasks (84.7\%) involve structured or image–text inputs, underscoring the benchmark’s emphasis on multimodal scientific comprehension.

Table~\ref{tab:dataset_statistics} details the scale and data sources for each task level. 
Lower-level tasks (L1--L2) focus on \textit{scientific literacy and information extraction}, while higher levels (L3--L5) involve \textit{reasoning, synthesis, and discovery}. 
Data are collected from a combination of public datasets (e.g., SuperGPQA, ScienceAgentBench) and self-curated scientific materials (e.g., arXiv and Nature papers), ensuring both reliability and diversity. 
Overall, HiSciBench includes {8,735 instances}, supporting comprehensive yet interpretable evaluation of scientific intelligence across multiple cognitive and modality dimensions.

\subsection{Design Principles}
\label{subsec:design_principles}

The architecture of HiSciBench is grounded in the recognition that scientific research expertise develops through distinct cognitive stages. Rather than treating scientific capability as a monolithic skill, we decompose it into a hierarchical framework that mirrors the natural progression of researchers: from acquiring basic scientific literacy, through mastering literature comprehension and analysis, to ultimately conducting original research and making novel discoveries.


Our benchmark encompasses five progressive levels, each representing a critical stage in the research lifecycle: \textbf{Level 1 (L1)} assesses fundamental scientific literacy; \textbf{Level 2 (L2)} evaluates information extraction from scientific literature; \textbf{Level 3 (L3)} tests comprehension and reasoning over multimodal research content; \textbf{Level 4 (L4)} examines synthesis and innovation through literature review generation; and \textbf{Level 5 (L5)} measures practical problem-solving through data-driven scientific discovery tasks.

\subsection{Task Hierarchy}
\label{subsec:task_hierarchy}

\textbf{L1: Scientific Literacy.}
\label{subsubsec:level1}

The foundation of scientific research capability rests upon a broad understanding of fundamental scientific concepts. Level 1 evaluates models' grasp of core knowledge across major scientific disciplines through carefully curated question-answering tasks.

\textbf{Task Description:} Models are presented with multiple-choice questions covering essential concepts in six major scientific domains: mathematics, physics, chemistry, astronomy, geography, and biology. These questions assess not merely factual recall, but conceptual understanding and the ability to apply fundamental principles to novel scenarios.

\textbf{Dataset Construction:} We construct the L1 dataset by systematically sampling from SuperGPQA, established benchmarks for scientific knowledge evaluation. For each of the six disciplines, we select 200 high-quality multiple-choice questions, ensuring balanced coverage of key topics within each domain. The resulting dataset comprises 1,200 question-answer pairs that collectively span the breadth of fundamental scientific knowledge expected of competent researchers.

\textbf{L2: Scientific Literature Parsing.}

\label{subsubsec:level2}
\textbf{Task Description:}  
Level 2 evaluates a model’s ability to extract, interpret, and translate information from multimodal scientific documents that contain dense combinations of text, equations, tables, and figures. It reflects the model’s competence in acquiring information from real-world research materials through two subtasks: \textbf{L2.1 Scientific Document Parsing} focuses on recognizing and reconstructing multimodal content from scientific pages, requiring accurate optical character recognition (OCR) and layout understanding to handle complex mathematical formulas, chemical structures, and specialized diagrams. \textbf{L2.2 Cross-lingual Scientific Translation} extends this by testing models’ ability to translate scientific texts across languages while preserving technical semantics, symbolic accuracy, and structural alignment, thereby supporting multilingual accessibility and global scientific communication.

\textbf{Dataset Construction:}  
For L2.1, we construct the dataset using LaTeX source files and their corresponding PDF documents collected from arXiv across multiple scientific disciplines. Each LaTeX source is converted to HTML and then to Markdown to preserve structural fidelity. The Markdown files are segmented according to page boundaries in the corresponding PDFs, and each page is rasterized into an image. This process yields a paired dataset of 629 samples encompassing mathematics (208), physics (357), astronomy (19), and geography (45), ensuring diversity in notation systems and visual conventions. 
For L2.2, we leverage the same Markdown sources from L2.1 and generate corresponding translations using Google Translate to create parallel multilingual data. This results in 629 paired translation samples with identical disciplinary distribution, enabling systematic evaluation of models’ cross-lingual understanding and translation accuracy in scientific contexts.

\textbf{L3: Literature Question Answering.}
\label{subsubsec:level3}

\textbf{Task Description:}  
Level 3 evaluates models’ capability to achieve deep comprehension and reasoning over multimodal scientific literature. Beyond surface-level parsing, this level requires integrating textual, visual, and symbolic information from research papers to answer content-grounded questions. It consists of two subtasks: \textbf{L3.1 Monolingual Literature QA}, which measures reading comprehension and reasoning within scientific documents in their original language, and \textbf{L3.2 Cross-lingual Literature QA}, which examines understanding and transfer across languages when queries and source materials differ linguistically. Together, these tasks assess whether models can connect heterogeneous modalities and maintain semantic consistency across linguistic contexts.

\textbf{Dataset Construction:}  
For L3.1, we construct a large-scale dataset by aggregating question–answer pairs from established benchmarks such as ScholarChemQA, ArXivQA, and PubMedQA, and augmenting them with additional samples derived from scientific papers collected via DOIs and official publication sources. Expert annotators define sub-domains for each discipline and extract the ten most frequent keywords from abstracts to map QA pairs to the most relevant scientific domain, ensuring precise annotation. The resulting dataset includes 5,514 QA pairs covering mathematics (821), physics (1,025), and other disciplines, each grounded in specific sections or figures of the source papers.  
For L3.2, we extend L3.1 to multilingual settings by leveraging the translated documents from Level 2. Using Qwen2.5-VL-72B, we automatically generate cross-lingual QA pairs, followed by manual verification of 20\% of the data to ensure factual and linguistic accuracy. This setup enables evaluation of models’ robustness in scientific reasoning under multilingual variation and cross-lingual information transfer.

\textbf{L4: Literature Review Generation.}
\label{subsubsec:level4}

The synthesis of existing research into coherent literature reviews represents a cornerstone of scientific writing, requiring both comprehensive understanding of a research area and the ability to identify patterns, gaps, and future directions. Level 4 evaluates models' capacity for producing high-quality scientific reviews.

\textbf{Task Description:} Given a research topic specified through keywords and a set of core papers, models must generate comprehensive literature reviews that synthesize current knowledge, identify research trends, critically analyze methodologies, and highlight open questions. This task demands not only comprehension of individual papers but sophisticated integration of ideas across multiple sources.

\textbf{Dataset Construction:} We compile review topics spanning six disciplines: mathematics, physics, chemistry, astronomy, geography and biology. For each discipline, we identify 10 recently published review paper titles from Nature and arXiv, ensuring coverage of diverse subfields and current research frontiers. 

\textbf{L5: Scientific Discovery.}
\label{subsubsec:level5}

The ultimate manifestation of scientific research capability lies in conducting original investigations and solving novel problems through data analysis and computational methods. Level 5 evaluates models' ability to translate scientific questions into executable solutions.

\textbf{Task Description:} Models are presented with scientific research problems requiring data-driven analysis, each accompanied by relevant datasets and domain-specific background knowledge. The task requires formulating appropriate computational approaches, implementing them in Python code, executing analyses, and interpreting results—mirroring the complete workflow of computational scientific research.

\textbf{Dataset Construction:} We adopt the ScienceAgentBench dataset, which provides carefully curated problems from four computation-intensive scientific disciplines. The dataset encompasses 67 distinct tasks, each comprising a problem description, associated data files, relevant domain knowledge, and expert-annotated reference solutions. Tasks span diverse analytical paradigms including statistical inference, numerical simulation, data visualization, and machine learning applications in scientific contexts.

\section{Experiments}

\subsection{\textbf{Experimental Setup}}
\label{subsec:exp_setup}



\subsubsection{\textbf{Evaluated Models}}
We evaluate a comprehensive collection of 18 state-of-the-art large language models spanning multiple categories: closed-source frontier models, open-source foundation models, vision-language architectures, and specialized research-oriented systems. This diverse model selection enables systematic assessment across the scientific literacy hierarchy, from basic knowledge understanding (L1) to advanced research capabilities (L5).

\noindent\textbf{Closed-source Multimodal Models.}
We include {GPT-5}\cite{openai2024gpt5}, a frontier multimodal foundation model supporting both text and visual inputs. GPT-5 serves as the primary reference for state-of-the-art performance across all levels of {HiSciBench}, participating in both text-input and vision-language evaluation tracks where applicable.

\noindent\textbf{Open-source Text-only Models.}
Our open-source text-only cohort includes nine models covering different parameter scales: \textit{i.Large-scale models} ($>$60B): {Llama-3.1-Instruct-70B} \cite{meta2024llama}, {DeepSeek-v3}\cite{deepseek2024v3}, {DeepSeek-R1} \cite{deepseek2024r1}; \textit{ii.Mid-scale models} (32B): {Qwen3-32B}\cite{alibaba2024qwen3}, {QWQ-32B}\cite{alibaba2024qwq}, {DeepSeek-R1-Distill-Qwen-32B (DeepSeek-R1-Distill-32B)}\cite{deepseek2024r1}
These models are evaluated on text-centric tasks, including {L1} (Scientific Literacy), {L2.2} (Cross-lingual Translation with text input), and text-based {L3} (Literature QA with extracted content).

\noindent\textbf{Open-source Vision-Language Models.}
We further evaluate five multimodal architectures for document-centric scientific tasks requiring visual understanding: \textit{i.Qwen-VL family:} {Qwen3-VL-8B}, {Qwen2.5-VL-7B}\cite{bai2025qwen2};\textit{ii.InternVL family:} {Intern-VL3-8B}, {Intern-VL3.5-8B}\cite{chen2024internvl}; \textit{iii.GLM-4.5V}\cite{vteam2025glm45vglm41vthinkingversatilemultimodal}
These models handle {L2.1} (OCR and Document Parsing), {L2.2} (Translation from visual inputs), and {L3} (Literature QA with PDF-based inputs), enabling evaluation of visual-textual grounding and multimodal reasoning in scientific contexts.

\noindent\textbf{Research-oriented Specialized Models.}
Finally, we include four systems explicitly designed for scientific research workflows:
\textit{i.Tongyi-DeepResearch}\cite{tongyidr}: a deep reasoning system optimized for multi-step research problem solving. \textit{ii.S1 family} (Orion Star): research-oriented large language models, including {S1-Base-Pro-32B\footnote{https://huggingface.co/ScienceOne-AI/S1-Base-32B}} (general scientific reasoning) and {S1-Literature} (literature comprehension and synthesis). \textit{iii.SurveyX}: a specialized system for automated literature review generation.
These systems primarily target higher-level reasoning tasks—{L4} (Literature Review Generation) and {L5} (Scientific Discovery)—representing the current frontier of AI-assisted scientific research and synthesis.


\subsubsection{\textbf{Evaluation Metrics}}
\label{subsec:metrics}

To ensure comprehensive and fair assessment across our hierarchical task structure, we employ evaluation metrics specifically tailored to each level's cognitive demands and output characteristics.

\textbf{L1: Scientific Literacy.} For multiple-choice questions testing fundamental knowledge, we employ classification accuracy as the primary metric. This straightforward measure effectively captures whether models possess the requisite scientific foundations.

\textbf{L2: Literature Parsing.} Document parsing and translation tasks necessitate evaluating the fidelity of generated structured representations against the ground truth. We use word-level accuracy to assess OCR performance (L2.1)\cite{liang2025improving} and the BLEU score to evaluate translation quality (L2.2)\cite{liang2024document}.

\textbf{L3: Literature Question Answering.} Similar to L1, we evaluate QA tasks using accuracy, as questions are designed with definitive answers. This metric directly measures whether models correctly comprehend and reason over scientific literature.

\textbf{L4: Literature Review.} Evaluating literature review quality requires multidimensional assessment, as reviews must satisfy both content quality and citation integrity requirements. We decompose evaluation into two primary dimensions:

\textbf{Content Quality.} Following the methodology established by SurveyX\cite{liang2025surveyx}, we employ LLM-as-Judge evaluation across five dimensions: (1) \textit{Coverage}:the comprehensiveness with which the review addresses the topic; (2) \textit{Structure}:the logical organization and coherence of presentation; (3) \textit{Relevance}:the degree to which content directly pertains to the topic; (4) \textit{Synthesis}:the effectiveness of integrating ideas across sources; and (5) \textit{Critical Analysis}:the depth of methodological critique and identification of research gaps. Each dimension is scored autonomously by advanced LLMs.

\textbf{Citation Quality.} Beyond content accuracy, scientific reviews must exhibit rigorous citation practices to ensure credibility and scholarly integrity. 
We evaluate citation quality from four complementary perspectives: 
(1) \textit{Verifiability}, which measures whether the cited references truly exist and whether their bibliographic information is accurate and properly formatted. 
This dimension includes metrics such as \textit{Verifiability Rate} and \textit{Metadata Accuracy}. 
(2) \textit{Coverage and Representativeness}, which captures the breadth and diversity of citations through the total \textit{Citation Count}, the number of \textit{Unique Sources}, and the \textit{Source Distribution Entropy} reflecting balance across publication venues. 
(3) \textit{Recency}, which quantifies the proportion of recently published papers among all citations, indicating the review’s awareness of the latest research progress. 
(4) \textit{Faithfulness}, which assesses whether each citation in the text accurately reflects the claims and findings of the original referenced work, ensuring that cited evidence is used in a truthful and contextually appropriate manner.

\textbf{L5: Scientific Discovery.} Evaluating code-based scientific problem solving requires metrics that reflect both functional correctness and overall code quality. We adopt the \textbf{Success Rate} (SR) metric, which assesses whether the generated code successfully fulfills the task-specific objectives. Beyond simple execution, SR measures whether the program’s outputs meet defined success criteria—such as achieving target performance on test data, producing correct predictions, or generating visualizations that accurately represent the underlying data. If the code fails to execute or produces invalid results, the SR is set to zero. This metric provides a direct and interpretable measure of task completion.

\subsection{\textbf{Overall Performance}}
\begin{figure}[t]
    \centering
    \includegraphics[width=0.68\linewidth]{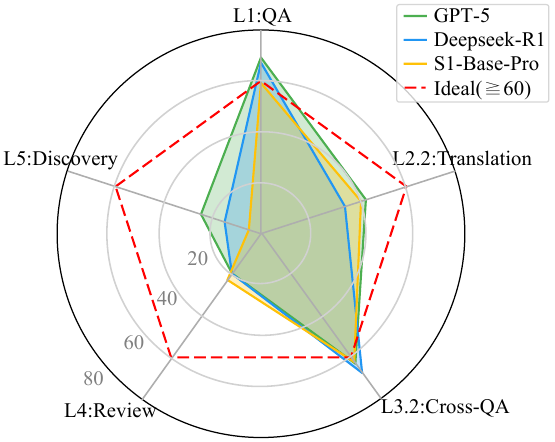}
    \caption{
    Radar chart of model performance across HiSciBench tasks (L1, L2.2, L3.2, L4, and L5). 
    GPT-5 demonstrates the most balanced performance across tasks, particularly excelling in reasoning (L3.2) and factual QA (L1). 
    Deepseek-r1 performs competitively in cross-lingual QA but trails in multimodal and discovery-oriented tasks. 
    S1 models are specialized for scientific research: {S1-Literature} is used for L4.1 (Literature Review Generation), 
    while {S1-Base-Pro (32B)} is applied to all other tasks. 
    The red dashed line indicates an \textit{ideal model} achieving at least 60 points across all tasks.
    }
    \label{fig:hiscibench_radar}
\end{figure}

Figure~\ref{fig:hiscibench_radar} present the overall performance of representative models across HiSciBench levels (L1--L5). 
Each level corresponds to an increasing cognitive complexity, ranging from factual recall (L1) to scientific discovery (L5). 
All results are averaged over six scientific disciplines, with a score of 0 indicating that a model does not support the corresponding task.

For the S1 family, two task-specific variants are used: {S1-Literature} for {L4.1 (Scientific Literature Review Generation)}, and {S1-Base-Pro (32B)} for all other levels (L1--L3, L5). 
This series is designed for \textit{fundamental scientific research}, emphasizing structured reasoning, domain-specific knowledge grounding, and controllable text generation.

Across all levels, {GPT-5} exhibits the strongest and most consistent performance, achieving the best or near-best scores in most tasks. 
{Deepseek-r1} performs competitively on text-based reasoning tasks, particularly in cross-lingual QA (L3.2) and verifiability (L4), but lacks multimodal adaptability. 
The {S1} models show clear domain specialization—S1-Literature performs well in literature synthesis (L4.1), while S1-Base-Pro demonstrates moderate generalization on other tasks. 
Overall, a notable performance gap remains between current systems and the ideal benchmark target ($\geq 60$\%), highlighting the challenge of achieving generalized scientific intelligence across modalities and reasoning levels.

\subsection{\textbf{Level-wise Detailed Results}}
\label{subsec:levelwise_results}

\textbf{L1: Scientific Literacy Assessment.}

\begin{table}[t]
\centering
\caption{Performance on L1: Scientific Literacy (Accuracy \%). Results are broken down by scientific discipline. AVG denotes the average over all disciplines.}
\label{tab:l1_results}
\resizebox{\linewidth}{!}{
\begin{tabular}{lccccccc}
\toprule
\textbf{Model} & \textbf{Math} & \textbf{Physics} & \textbf{Chemistry} & \textbf{Astronomy} & \textbf{Geography} & \textbf{Biology} & \textbf{AVG} \\
\midrule
GPT-5 & \textbf{84.50} & \textbf{70.50} & \underline{65.50} & \textbf{66.00} & \underline{66.00} & \textbf{62.50} & \textbf{69.17} \\
Deepseek-r1 & \underline{84.00} & \underline{68.50} & \textbf{66.00} & \textbf{66.00} & 61.00 & 57.50 & \underline{67.17} \\
Deepseek-v3 & 63.50 & 68.00 & \textbf{66.00} & \underline{64.00} & \textbf{69.50} & \underline{60.00} & 65.17 \\
Llama-3.1-70B & 32.50 & 38.00 & 32.50 & 35.50 & 35.00 & 38.00 & 35.25 \\
\midrule
DeepSeek-R1-Distill-32B& 46.00 & 34.00 & 34.00 & 40.00 & 39.50 & 38.00 & 38.58 \\
Qwen3-32B & 71.50 & 63.50 & 56.60 & 56.50 & 49.00 & 50.00 & 57.85 \\
QWQ-32B & 71.00 & 54.50 & 46.50 & 50.00 & 47.50 & 47.50 & 52.83 \\
S1-Base-Pro-32B & 70.00 & 62.50 & 61.50 & 58.50 & 56.00 & 46.50 & 59.17 \\
\midrule
Tongyi-DeepResearch & 67.00 & 50.00 & 45.00 & 42.00 & 43.00 & 52.00 & 49.83 \\
\bottomrule
\end{tabular}}
\end{table}

Table~\ref{tab:l1_results} presents performance on L1, evaluating factual recall and conceptual understanding across six scientific disciplines as the foundation of HiSciBench.
{GPT-5} achieves the highest average accuracy (69.17\%), excelling in mathematics (84.50\%) and physics (70.50\%), demonstrating that large-scale multimodal pretraining effectively captures symbolic and conceptual scientific patterns.
{Deepseek-r1} ranks second (67.17\%) with competitive chemistry and astronomy results but weaker biology and geography performance.
Among 32B-scale open models, {S1-Base-Pro-32B} leads (59.17\%), outperforming Qwen3-32B (57.85\%) and substantially exceeding DeepSeek-R1-Distill-32B (38.58\%), whose 28.6-point drop from the full model (67.17\%) reveals that aggressive compression severely erodes factual completeness and reasoning fidelity.
The retrieval-oriented Tongyi-DeepResearch (49.83\%) shows limited generalization beyond its specialized focus.

Performance varies considerably across disciplines: mathematics achieves the highest accuracy (65.56\%), while biology remains most challenging (50.22\%) due to greater linguistic and conceptual variability.
General-purpose models consistently outperform specialized agents by 10.4 percentage points, emphasizing the importance of large-scale factual priors.
However, even GPT-5's 70\% accuracy indicates current LLMs still fall short of human-level scientific reasoning.

These findings reveal that scientific comprehension scales with model size and multimodal pretraining, yet achieving human-comparable literacy remains an open challenge.


\begin{table}[t]
\centering
\caption{
Performance on \textbf{L2: Scientific Literature Parsing and Translation} (BLEU Score). 
L2.1 evaluates the ability to parse PDF figures into structured Markdown (Document Parsing), 
while L2.2 measures cross-lingual scientific translation from English to Chinese with terminology preservation. 
}
\label{tab:l2_results}
\resizebox{\linewidth}{!}{
\begin{tabular}{cccccc}
\toprule
\textbf{Model} & \textbf{Math} & \textbf{Physics} & \textbf{Astronomy} & \textbf{Biology} & \textbf{Average} \\
\midrule
\multicolumn{6}{l}{\cellcolor{green!10}\textbf{L2.1: Scientific Document Parsing (Vision-Language Input)}} \\
GPT-5 & \textbf{52.75} & \underline{74.30} & \underline{70.82} & \textbf{72.57} & \textbf{67.61} \\
Qwen3-VL-8B & 41.43 & \textbf{74.90} & \textbf{71.85} & \underline{70.85} & 64.76 \\
Qwen2.5-VL-7B & \underline{49.32} & 71.44 & 68.47 & 69.97 & 64.80 \\
Intern-VL3.5-8B & 7.26 & 15.11 & 3.92 & 11.81 & 9.53 \\
Intern-VL3-8B & 6.53 & 11.96 & 3.44 & 7.80 & 7.43 \\
\midrule
\multicolumn{6}{l}{\cellcolor{green!10}\textbf{L2.2: Cross-lingual Scientific Translation (Text Input)}} \\
GPT-5 & 37.47 & \textbf{41.04} & \textbf{45.21} & \textbf{49.45} & \textbf{43.29} \\
Deepseek-v3 & \underline{38.94} & \underline{37.57} & 36.20 & 43.20 & 38.98 \\
Tongyi-DeepResearch & 36.20 & 32.28 & 41.16 & 41.70 & 37.84 \\
Deepseek-r1  & 36.25 & 32.35 & 34.08 & 35.97 & 34.66 \\
S1-base-Pro-32B & \textbf{41.49} & 36.67 & \underline{41.66} & \underline{45.31} & \underline{41.28} \\
\midrule
\multicolumn{6}{l}{\cellcolor{green!10}\textbf{L2.2: Cross-lingual Scientific Translation (Vision-Language Input)}} \\
GPT-5 & \underline{21.62} & \textbf{28.31} & \textbf{28.23} & \textbf{36.50} & \textbf{28.67} \\
Qwen3-VL-8B & \textbf{24.74} & \underline{26.27} & \underline{26.51} & \underline{35.87} & \underline{28.35} \\
Qwen2.5-VL-7B & 16.42 & 15.90 & 9.06 & 24.37 & 16.44 \\
Intern-VL3.5-8B & 4.36 & 4.36 & 1.43 & 7.48 & 4.41 \\
Intern-VL3-8B & 4.49 & 4.79 & 1.55 & 4.97 & 3.95 \\
\bottomrule
\end{tabular}
}
\end{table}

\textbf{L2: Scientific Document Parsing.}

Table~\ref{tab:l2_results} and Figure~\ref{fig:l2_radar} report results on \textbf{L2: Scientific Literature Parsing and Translation}, which evaluate a model’s ability to \textit{read}, \textit{understand}, and \textit{translate} scientific documents across modalities. 
L2.1 examines structured document parsing from PDFs (vision-language input), while L2.2 measures cross-lingual translation accuracy under text and visual inputs.

Across both tasks, {GPT-5} consistently achieves the best performance, with BLEU scores of {67.61} on document parsing and {43.29} on translation. 
In L2.1, GPT-5 and Qwen3-VL-8B perform comparably in physics and astronomy, suggesting strong visual parsing capabilities. 
By contrast, smaller models such as Intern-VL3 and VL3.5 degrade sharply (BLEU $<$ 10), underscoring the importance of large-scale multimodal pretraining for robust text–vision alignment and formula extraction.
In L2.2, GPT-5 again leads, followed by S1-base-Pro (41.28) and Deepseek-v3 (38.98). 
S1-base-Pro benefits from scientific text fine-tuning, achieving strong terminology preservation but lower fluency than GPT-5. 
Text-only models consistently outperform their vision-based counterparts, indicating that visual inputs introduce noise into linguistic reasoning.

Overall, these findings reveal two key insights:  
(1) Large multimodal models like GPT-5 excel in both document understanding and cross-lingual reasoning;  
(2) Cross-modal consistency remains a major bottleneck—current vision-language systems fail to match the semantic precision of text-only models.  
Bridging this gap will require more effective visual–text fusion and pretraining strategies explicitly optimized for document-level semantic preservation.

\begin{table*}[t]
\centering
\caption{
Performance on \textbf{L3: Scientific Literature Question Answering (QA)} (Accuracy \%). 
L3.1 measures monolingual reasoning, while L3.2 evaluates cross-lingual QA with English documents and Chinese questions. 
Each task includes both Vision-Language and Text-only input settings.
}
\label{tab:l3_results_combined}
\resizebox{.85\linewidth}{!}{
\begin{tabular}{lccccccc}
\toprule
\textbf{Task / Model} & \textbf{Math} & \textbf{Physics} & \textbf{Chemistry} & \textbf{Astronomy} & \textbf{Geography} & \textbf{Biology} & \textbf{Average} \\
\midrule
\multicolumn{8}{l}{\cellcolor{green!10}\textbf{L3.1: Monolingual Literature QA (Vision-Language Input, Full-text / Fragment-based)}} \\
GPT-5 & 59.30/71.16 & 61.00/67.43 & 69.07/71.49 & 60.61/62.42 & 69.20/73.20 & 90.68/89.93 & 73.39/76.75 \\
GLM-4.5V & 65.33/72.58 & 66.80/75.05 & 74.49/79.41 & 62.42/72.73 & 67.20/75.20 & 89.24/89.93 & 75.62/80.45 \\
Qwen3VL-30B-A3B & 62.56/69.50 & 63.40/68.57 & 72.22/72.40 & 67.88/73.33 & 66.40/70.00 & 88.32/87.26 & 73.98/76.28 \\
Qwen3-VL-8B & 57.54/66.43 & 63.80/70.10 & 73.14/76.24 & 63.64/72.73 & 66.00/70.80 & 86.78/85.92 & 72.80/76.28 \\
Qwen2.5-VL-7B & 50.75/63.12 & 54.60/63.62 & 65.46/66.06 & 60.61/62.42 & 58.40/65.20 & 83.20/78.62 & 66.73/69.26 \\
Intern-VL3.5-8B & 44.22/66.90 & 47.20/65.90 & 49.21/68.10 & 52.73/63.03 & 51.20/72.00 & 50.72/80.58 & 49.05/71.92 \\
Intern-VL3-8B & 55.03/64.30 & 52.00/64.38 & 58.92/68.10 & 51.52/67.88 & 62.40/62.40 & 83.30/78.01 & 65.67/69.76 \\
Intern-VL3.5-38B & 50.50/43.26 & 49.00/43.24 & 51.69/40.95 & 53.33/46.67 & 61.60/40.95 & 75.31/53.44 & 60.47/47.77 \\
\midrule
\multicolumn{8}{l}{\cellcolor{green!10}\textbf{L3.2: Cross-lingual Literature QA (Vision-Language Input)}} \\
GPT-5 & \textbf{80.00} & \textbf{92.00} & -- & \textbf{84.21} & -- & \textbf{88.89} & \textbf{86.28} \\
Qwen3-VL-8B & \textbf{80.00} & \underline{82.00} & -- & \underline{68.42} & -- & \textbf{88.89} & \underline{79.83} \\
Qwen2.5-VL-7B & \underline{68.00} & 78.00 & -- & \underline{68.42} & -- & \underline{75.56} & 72.50 \\
Intern-VL3.5-8B & 36.00 & 28.00 & -- & 26.32 & -- & 20.00 & 27.58 \\
Intern-VL3-8B & 28.00 & 26.00 & -- & 36.84 & -- & 17.78 & 27.16 \\

\midrule

\multicolumn{8}{l}{\cellcolor{green!10}\textbf{L3.1: Monolingual Literature QA (Text Input)}} \\
Deepseek-v3 & \underline{86.21} & \underline{92.86} & \textbf{92.31} & -- & \underline{85.29} & \textbf{97.88} & \textbf{96.20} \\
Deepseek-r1 & 82.76 & \underline{92.86} & 83.08 & -- & \underline{85.29} & \underline{96.02} & \underline{93.43} \\
DeepSeek-R1-Distill-32B & 68.97 & 85.71 & 78.46 & -- & \textbf{88.24} & 87.12 & 85.42 \\
Tongyi-DeepResearch & 82.76 & \textbf{96.43} & 74.62 & -- & \textbf{88.24} & 88.31 & 86.55 \\
S1-Base-Pro-32B & \textbf{93.10} & \underline{92.86} & \underline{86.15} & -- & 82.35 & 93.36 & 91.00 \\
S1-Base-8B & 72.41 & \underline{92.86} & 37.69 & -- & 44.12 & 40.50 & 42.71 \\
\midrule
\multicolumn{8}{l}{\cellcolor{green!10}\textbf{L3.2: Cross-lingual Literature QA (Text Input)}} \\
GPT-5 & 62.00 & \textbf{66.00} & -- & \underline{68.42} & -- & \underline{55.56} & 63.00 \\
Deepseek-r1 & \textbf{66.00} & \textbf{66.00} & -- & \textbf{73.68} & -- & \textbf{64.44} & \textbf{67.53} \\
Deepseek-v3 & 52.00 & \underline{62.00} & -- & \textbf{73.68} & -- & \underline{55.56} & 60.81 \\
Tongyi-DeepResearch & \underline{64.00} & \underline{62.00} & -- & \underline{68.42} & -- & \textbf{64.44} & \underline{64.72} \\
S1-Base-Pro-32B & 60.00 & \underline{62.00} & -- & \textbf{73.68} & -- & 53.33 & 62.25 \\
\bottomrule
\end{tabular}
}
\end{table*}

\noindent\textbf{L3: Scientific Literature QA.}

Table~\ref{tab:l3_results_combined} summarizes model performance on \textbf{L3: Scientific Literature QA}, which evaluates scientific reasoning and comprehension at the document level. 
L3.1 measures monolingual reasoning within English scientific papers, while L3.2 tests cross-lingual understanding where questions are posed in Chinese but source documents remain in English. 
These tasks require models to perform contextual inference, information integration, and domain-aware grounding beyond simple factual recall.

\noindent\textbf{ L3.1 Monolingual Literature QA.} 
Across both text-only and vision-language settings, large-scale foundation models exhibit strong comprehension capabilities. 
Among text-based systems, {Deepseek-v3} achieves the highest accuracy ({96.20\%}), closely followed by {Deepseek-r1} (93.43\%) and {S1-Base-Pro-32B} (91.00\%). 
These results indicate that monolingual reasoning benefits substantially from large language model pretraining and scientific corpus alignment. 
By contrast, smaller specialized agents such as S1-Base-8B (42.71\%) or distilled variants show limited transfer ability, suggesting that scale and retained factual priors are key for document-level consistency.

In the vision-language configuration, {GLM-4.5V} and {GPT-5} achieve the best performance (average {80.45\%} and {76.75\%}, respectively), demonstrating robust multimodal understanding across text–figure integration. 
However, a noticeable accuracy gap remains between fragment-based and full-text reasoning, implying that document segmentation can simplify inference by focusing attention on semantically dense regions. 
Performance drops sharply in smaller visual models (e.g., Intern-VL3/3.5), revealing their weak cross-modal grounding and limited ability to parse scientific figures and equations.

\noindent\textbf{L3.2 Cross-lingual Literature QA.} 
This task evaluates whether models can align multilingual semantics with scientific context. 
{GPT-5} again achieves the best results in both text ({63.00\%}) and vision-language ({86.28\%}) settings, confirming strong cross-lingual reasoning and terminology transfer. 
{Deepseek-r1} (67.53\%) also perform competitively, highlighting that bilingual fine-tuning and instruction-level alignment can narrow the gap with general-purpose multilingual models. 
Smaller multimodal models (Intern-VL3/3.5) struggle significantly, often failing to map cross-language entity correspondences.


In summary, L3 tasks demonstrate that scientific comprehension requires deep integration of factual memory, contextual inference, and multilingual grounding. 
While frontier models like GPT-5 and Deepseek-v3 approach human-level monolingual understanding, sustaining comparable performance in cross-lingual and multimodal contexts remains a key frontier for next-generation research models.

\textbf{L4: Literature Review Generation.}
\label{subsubsec:results_l4}

\begin{table*}[t]
\centering
\caption{
Performance on \textbf{L4: Literature Review Evaluation}.
Upper section: \textit{Content Quality} (Score 1–5) evaluated via LLM-as-Judge.  
Lower section: \textit{Citation Quality} assessed by factual and bibliographic accuracy.
}
\label{tab:l4_combined_quality}
\resizebox{0.9\linewidth}{!}{
\begin{tabular}{lcccccc}
\toprule
\textbf{Metric} & \textbf{GPT-5} & \textbf{Deepseek-r1} & \textbf{Deepseek-v3} & \textbf{Tongyi-DeepResearch} & \textbf{SurveyX} & \textbf{S1-Literature} \\
\midrule
\rowcolor{green!10}
\multicolumn{7}{l}{\textit{Content Quality (Score 1–5)}} \\
Coverage & \textbf{5.00} & 4.97 & 4.35 & 4.97 & \underline{4.98} & 4.85 \\
Structure & \textbf{5.00} & \underline{4.97} & 4.93 & 4.97 & 4.45 & 4.83 \\
Relevance & \textbf{5.00} & \underline{4.97} & 4.93 & 4.97 & 4.88 & 4.80 \\
Synthesis & \textbf{5.00} & 4.32 & 4.03 & \underline{4.92} & 4.33 & 4.83 \\
Critical Analysis & \textbf{4.95} & 4.25 & 3.97 & \underline{4.88} & 4.42 & 4.77 \\
Overall Score & \textbf{4.99} & 4.65 & 4.44 & \underline{4.94} & 4.61 & 4.82 \\
\midrule
\rowcolor{green!10}
\multicolumn{7}{l}{\textit{Citation Quality (Accuracy and Reliability Metrics)}} \\
Verifiability Rate (\%) & 19.3 & 19.4 & 17.9 & -- & \textbf{71.4} & \underline{22.4} \\
Metadata Accuracy (\%) & 2.6 & 3.4 & 2.1 & -- & \textbf{45.6} & \underline{11.5} \\
Faithfulness Rate (\%) & \underline{10.5} & 4.5 & 4.8 & -- & \textbf{27.2} & 8.0 \\
Citation Count & 35.0 & 21.4 & 11.4 & -- & \textbf{86.6} & \underline{81.0} \\
Source Count & 2.6 & 2.3 & 1.3 & -- & \underline{8.8} & \textbf{11.1} \\
Source Entropy & 0.63 & 0.64 & 0.33 & -- & \underline{0.88} & \textbf{1.96} \\
Recency Rate (\%) & 0.8 & 0.9 & 1.0 & -- & \textbf{16.1} & \underline{4.3} \\
\midrule
\rowcolor{green!10}
\multicolumn{7}{l}{\textit{Output Token}} \\
Average Length (tokens) & 7,578 & 2,349 & 1,482 & 5,338 & 22,253 & 39,098 \\
\bottomrule
\end{tabular}
}
\end{table*}

Table~\ref{tab:l4_combined_quality} summarizes performance on L4, covering content and citation quality across multiple dimensions.
\textbf{Content quality} is exceptionally high: GPT-5 achieves a near-perfect Overall Score (4.99/5.0), with Tongyi-DeepResearch closely following (4.94).
However, \textbf{Critical Analysis} remains the weakest dimension across all models (3.97–4.93), with Deepseek models particularly struggling in both Synthesis (4.03–4.32) and Critical Analysis, indicating that generating sophisticated analytical perspectives remains challenging for current LLMs.

A striking gap emerges in \textbf{citation reliability}.
Specialized systems significantly outperform general models: SurveyX leads with 71.4\% Verifiability Rate and 45.6\% Metadata Accuracy, vastly exceeding GPT-5's 19.3\% and 2.6\%, respectively.
General models exhibit severe citation hallucination, with verifiability rates below 20\% (GPT-5: 19.3\%, Deepseek-v3: 17.9\%), generating non-existent or unverifiable references despite producing fluent content.
These findings highlight that Retrieval-Augmented Generation architectures are crucial for grounding citations to authentic sources, whereas current general-purpose LLMs prioritize surface coherence over factual traceability.

\textbf{L5: Scientific Discovery and Problem Solving.}
\label{subsubsec:results_l5}

Table~\ref{tab:l5_results} presents the code generation and scientific problem-solving capabilities measured by task success rate.
As shown in Table~\ref{tab:l5_results}, all models achieve relatively low success rates, indicating the difficulty of data-driven scientific reasoning. GPT-5 achieves the best overall performance (24.75\%), followed by Deepseek-r1 (21.05\%), while other models remain below 15\%. Across disciplines, Geography yields the highest accuracy (33.33\%), and Chemistry remains the most challenging (15.00\%).
Although Deepseek-r1 shows stronger performance in Biology and Chemistry, GPT-5 demonstrates more balanced generalization. Smaller or distilled models (e.g., Deepseek-v3, QWQ-32B) suffer severe degradation, highlighting the dependence on model scale. Overall, results suggest that current models can execute structured reasoning but still lack the inductive and hypothesis-forming abilities required for genuine scientific discovery.

\begin{table}[t]
\centering
\caption{Performance on L5: Scientific Discovery (Success Rate \%). Task completion rates across disciplines.}
\label{tab:l5_results}
\resizebox{.85\linewidth}{!}{
\begin{tabular}{lcccc}
\toprule
\textbf{Model} & \textbf{Chemistry} & \textbf{Geography} & \textbf{Biology} & \textbf{Average} \\
\midrule
GPT-5 & \underline{15.00} & \textbf{33.33} & \underline{25.93} & \textbf{24.75} \\
Deepseek-r1 & \textbf{15.00} & \underline{18.52} & \textbf{29.63} & \underline{21.05} \\
Deepseek-v3 & 5.00 & 14.81 & 22.22 & 14.01 \\
QWQ-32B & 5.00 & 3.70 & 14.81 & 7.84 \\
DeepSeek-R1-Distill-32B & 0.00 & 11.11 & 7.41 & 6.17 \\
Llama-3.1-70B & 5.00 & 7.41 & 3.70 & 5.37 \\
S1-Base-Pro-32B & 5.00 & 11.11 & 11.11 & 9.07 \\
\bottomrule
\end{tabular}
}
\end{table}

\subsection{\textbf{Analysis and Discussions}}
\label{subsec:analysis}



\begin{figure}[t]
    \centering
    \includegraphics[width=0.9\linewidth]{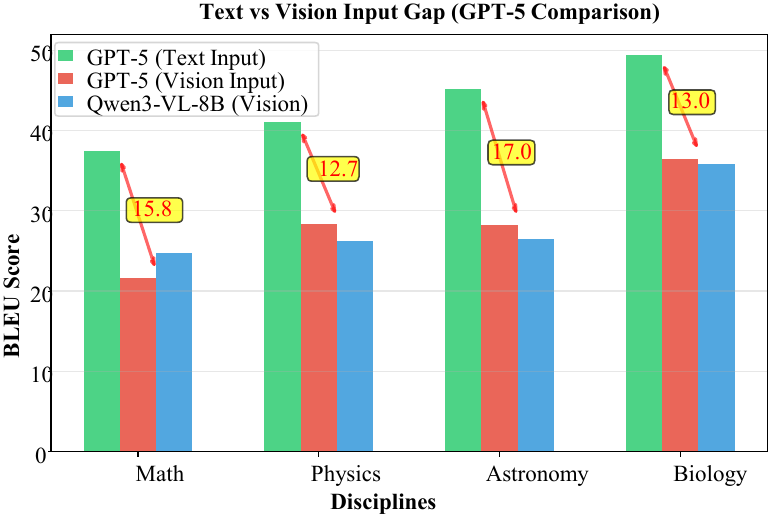}
    \caption{
    Radar comparison of GPT-5’s performance on \textbf{L2.2: Cross-lingual Scientific Translation} across four disciplines under \textit{text-only} and \textit{vision-language} inputs. 
    The text-input curve achieves higher BLEU scores (37–49), while the vision-language curve consistently trails by 12–17 BLEU, revealing that visual context currently introduces semantic noise instead of aiding linguistic reasoning.
    }
    \label{fig:l2_radar}
\end{figure}

\noindent\textbf{Q1.Does modality affect performance?}
Figure~\ref{fig:l2_radar} compares GPT-5’s cross-lingual translation performance across four scientific disciplines (Mathematics, Physics, Astronomy, and Biology) under \textit{text-only} and \textit{vision-language} input modalities. 
A consistent degradation is observed when visual information is introduced alongside text, indicating that multimodal inputs can interfere with semantic translation fidelity rather than enhance it.

\begin{figure}[t]
    \centering
    \includegraphics[width=0.9\linewidth]{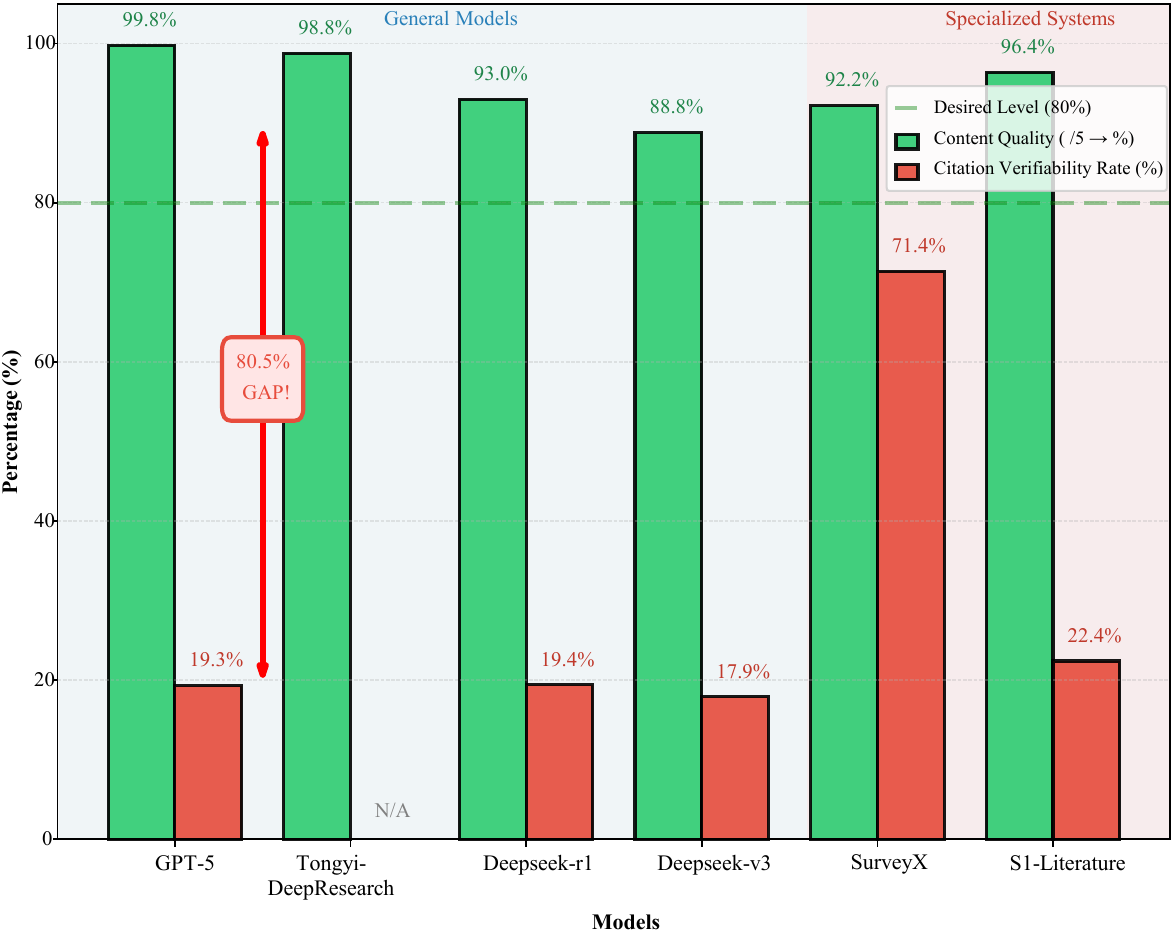}
    \caption{
    Comparison of \textbf{content quality} and \textbf{citation verifiability rate} across general-purpose and specialized models on \textbf{L4: Scientific Literature Review Generation}.
    While all models achieve near-perfect content quality (88.8–99.8\%), their citation verifiability remains drastically lower (17–22\%), revealing an \textbf{80\% factuality gap}.
    This demonstrates that LLMs can generate coherent and well-structured scientific reviews, yet often fail to provide verifiable or authentic citations.
    The results suggest that \textit{surface-level fluency does not equate to factual grounding}, emphasizing the need for stronger citation retrieval and source alignment mechanisms in future scientific LLMs.
    }
    \label{fig:l4_quality_gap}
\end{figure}

\noindent\textbf{Q2. Can LLMs judge their own scientific quality?}
Figure~\ref{fig:l4_quality_gap} reveals a striking disparity between self-assessed content quality and citation reliability in L4.
LLM-as-Judge evaluations show models achieve exceptionally high content scores (88.8–99.8\%), consistently surpassing the 80\% benchmark.
However, external verification exposes a severe \textit{80\% factuality gap}: verifiability rates plummet to 17–22\%.
While models generate fluent, well-structured reviews, their references lack grounding in authentic sources, indicating systematic overestimation of scientific coherence.
Notably, even specialized systems (e.g., S1-Literature: 96.4\% content quality, 22.4\% citation validity) exhibit this tendency, demonstrating that domain specialization alone cannot guarantee factual reliability.
These findings highlight a fundamental limitation: \textit{surface-level fluency does not equate to epistemic accuracy}.
Closing this gap requires incorporating citation grounding, retrieval-based verification, and source alignment rather than relying solely on generative confidence.

\noindent\textbf{Q3. Why do LLMs produce high-quality reviews but unreliable citations?}
Qualitative analysis of L4 reveals five recurring failure modes explaining the gap between content quality and citation reliability in Table~\ref{tab:l4_combined_quality}.
\textit{Citation fabrication} accounts for 81\% of unverifiable references in general models, generating realistic-looking but non-existent citations that critically undermine scientific credibility.
\textit{Metadata inaccuracy} affects 97\% of citations, with models frequently misreporting authors, years, or journal names even when papers exist, resulting in false traceability.
\textit{Recency lag} limits models to 0–1\% recent citations, failing to reflect latest developments in rapidly evolving fields.
\textit{Source concentration} manifests as repeated citation of a small cluster (average 5.3 sources) rather than broad literature sampling, causing insufficient coverage and evidence redundancy.
\textit{Faithfulness deficit} produces an 89\% unfaithfulness rate, where generated claims diverge from actual source content through misattributed findings, overstated conclusions, or out-of-context citations.

\noindent\textbf{Q4.When Does Scientific Discovery Fail?} 
Table~\ref{tab:l5_errors} presents two representative failure modes at L5, where models must translate scientific instructions into executable workflows reflecting domain logic. 
In the first case (Incomplete Solution), the model correctly processes demographic data and generates visualizations but produces three separate subplots instead of a composite map, revealing deficient \textit{conceptual synthesis} in integrating spatial and demographic layers. 
In the second case (Code Logic Error), the model attempts to count five-year warm spells but reduces the array along the latitude axis rather than the temporal dimension, yielding a zero-valued map. 
Though syntactically correct, the code lacks \textit{spatiotemporal reasoning} consistency. 
These failures demonstrate that L5 errors stem not from coding mistakes but from misalignment between scientific intent and computational reasoning, exposing the gap between code generation and genuine scientific understanding.
\begin{table}[t]
\label{tab:l5_errors}
\centering
\caption{Representative error cases in \textbf{L5: Scientific Discovery}. Each illustrates distinct weaknesses in data reasoning or code execution.}
\small
\begin{tabular}{p{0.95\linewidth}}
\hline
\rowcolor{green!10}
\textbf{Error Type: Incomplete Solution}\\
\textbf{Task:} “Map existing bus stops’ service area with enriched map layers for different census blocks…” (Hamilton County, TN).\\
\textbf{Model Output (excerpt):} \texttt{pred\_transit\_access.py} builds three subplots via \texttt{plt.subplots(1, 3, ...)} and calls \texttt{plot\_metric(...)} for population density, poverty rate, and no-vehicle rate separately.\\
\textbf{What went wrong:} The instructions expect a single composite map with layered overlays. By plotting metrics in independent panels, the model fails to merge demographic layers with the transit boundary, leading to an incomplete submission despite correct preprocessing.\\
\textbf{Correct reasoning:} Render a unified map with stacked layers: use the demographic polygon base, apply graduated symbology, and overlay the bus service region and coverage mask within one axes object to match the reference visualization.\\
\hline
\rowcolor{green!10}
\textbf{Error Type: Code Logic Error}\\
\textbf{Task:} “Use 240 years of annual surface temperature to count five-year warm spells exceeding $280\,\mathrm{K}$ and plot the map.”\\
\textbf{Model Output (excerpt):} The code uses \texttt{sliding\_window\_view(..., axis=0)} but reduces via \texttt{windows.all(axis=1)}, producing a uniform zero-valued map.\\
\textbf{What went wrong:} Reducing along \texttt{axis=1} collapses the latitude dimension instead of the run-length axis, effectively demanding all latitudes exceed $280\,\mathrm{K}$ simultaneously.\\
\textbf{Correct reasoning:} Reduce over the trailing run-length axis—e.g., \texttt{windows.all(axis=-1)}—so that each grid cell independently counts consecutive warm years, restoring the expected spatial gradient.\\
\hline
\end{tabular}
\end{table}

\section{Conclusion}

We introduced {HiSciBench}, a hierarchical benchmark for evaluating scientific intelligence in large language models (LLMs) across six disciplines and five cognitive levels (L1–L5). 
Built from authentic scientific literature, HiSciBench captures the full research workflow—from factual recall to hypothesis generation—enabling systematic, interpretable assessment of scientific reasoning.
Experiments show that current LLMs excel at factual and linguistic understanding (L1–L2) but struggle with higher-order cognition. 
Performance declines markedly in multimodal and cross-lingual tasks (L2–L3), citation faithfulness in literature synthesis (L4), and scientific code reasoning (L5). 
These results reveal persistent weaknesses in epistemic grounding, multimodal integration, and procedural reasoning.
HiSciBench thus provides a unified framework for diagnosing such limitations and guiding the development of next-generation scientific AI—models capable of reliable understanding, faithful synthesis, and verifiable discovery.


\bibliographystyle{IEEEtran}
\bibliography{references}

@inproceedings{liang2025improving,
  title={Improving MLLM’s Document Image Machine Translation via Synchronously Self-reviewing Its OCR Proficiency},
  author={Liang, Yupu and Zhang, Yaping and Zhang, Zhiyang and Chen, Zhiyuan and Zhao, Yang and Xiang, Lu and Zong, Chengqing and Zhou, Yu},
  booktitle={Findings of the Association for Computational Linguistics: ACL 2025},
  pages={23659--23678},
  year={2025}
}

@inproceedings{liang2024document,
  title={Document image machine translation with dynamic multi-pre-trained models assembling},
  author={Liang, Yupu and Zhang, Yaping and Ma, Cong and Zhang, Zhiyang and Zhao, Yang and Xiang, Lu and Zong, Chengqing and Zhou, Yu},
  booktitle={Proceedings of the 2024 Conference of the North American Chapter of the Association for Computational Linguistics: Human Language Technologies (Volume 1: Long Papers)},
  pages={7084--7095},
  year={2024}
}

@inproceedings{lu2024mathvista,
  title={MathVista: Evaluating Mathematical Reasoning of Foundation Models in Visual Contexts},
  author={Lu, Pan and Bansal, Hritik and Xia, Tony and Liu, Jiacheng and Li, Chunyuan and Hajishirzi, Hannaneh and Cheng, Hao and Chang, Kai-Wei and Galley, Michel and Gao, Jianfeng},
  booktitle={The Twelfth International Conference on Learning Representations},
  year={2024}
}

@inproceedings{cai2025sciassess,
  title={Sciassess: Benchmarking llm proficiency in scientific literature analysis},
  author={Cai, Hengxing and Cai, Xiaochen and Chang, Junhan and Li, Sihang and Yao, Lin and Changxin, Wang and Gao, Zhifeng and Wang, Hongshuai and Yongge, Li and Lin, Mujie and others},
  booktitle={Findings of the Association for Computational Linguistics: NAACL 2025},
  pages={2335--2357},
  year={2025}
}

@article{mirza2024large,
  title={Are large language models superhuman chemists?},
  author={Mirza, Adrian and Alampara, Nawaf and Kunchapu, Sreekanth and R{\'\i}os-Garc{\'\i}a, Marti{\~n}o and Emoekabu, Benedict and Krishnan, Aswanth and Gupta, Tanya and Schilling-Wilhelmi, Mara and Okereke, Macjonathan and Aneesh, Anagha and others},
  journal={arXiv preprint arXiv:2404.01475},
  year={2024}
}

@inproceedings{singh2023scirepeval,
  title={Scirepeval: A multi-format benchmark for scientific document representations},
  author={Singh, Amanpreet and D’Arcy, Mike and Cohan, Arman and Downey, Doug and Feldman, Sergey},
  booktitle={Proceedings of the 2023 Conference on Empirical Methods in Natural Language Processing},
  pages={5548--5566},
  year={2023}
}

@article{lala2023paperqa,
  title={PaperQA: Retrieval-Augmented Generative Agent for Scientific Research},
  author={Lala, Jakub and O'Donoghue, Odhran and Shtedritski, Aleksandar and Cox, Sam and Rodriques, Samuel G and White, Andrew D},
  journal={arXiv preprint arXiv:2312.07559},
  year={2023}
}

@inproceedings{noh-etal-2025-scholarbench,
    title = "{S}cholar{B}ench: A Bilingual Benchmark for Abstraction, Comprehension, and Reasoning Evaluation in Academic Contexts",
    author = "Noh, Dongwon  and
      Koh, Donghyeok  and
      Yuk, Junghun  and
      Kim, Gyuwan  and
      Lee, Jae Yong  and
      Lim, KyungTae  and
      Park, Cheoneum",
    editor = "Christodoulopoulos, Christos  and
      Chakraborty, Tanmoy  and
      Rose, Carolyn  and
      Peng, Violet",
    booktitle = "Findings of the Association for Computational Linguistics: EMNLP 2025",
    month = nov,
    year = "2025",
    address = "Suzhou, China",
    publisher = "Association for Computational Linguistics",
    url = "https://aclanthology.org/2025.findings-emnlp.465/",
    pages = "8750--8782",
    ISBN = "979-8-89176-335-7"
}

@article{openai2024gpt5,
  title={GPT-5 Technical Report},
  author={OpenAI},
  journal={Technical Report},
  year={2024}
}

@article{meta2024llama,
  title={Llama 3: Open Foundation and Fine-Tuned Chat Models},
  author={Meta AI},
  journal={Technical Report},
  year={2024}
}

@article{deepseek2024r1,
  title={DeepSeek-R1: Reasoning-First Approach to Large Language Model},
  author={DeepSeek AI},
  journal={arXiv preprint},
  year={2024}
}

@article{deepseek2024v3,
  title={DeepSeek-V3: Towards AGI with Efficient Training},
  author={DeepSeek AI},
  journal={arXiv preprint},
  year={2024}
}

@article{alibaba2024qwq,
  title={QWQ: Question-Weighted Query Model for Mathematical Reasoning},
  author={Alibaba DAMO Academy},
  journal={Technical Report},
  year={2024}
}

@article{alibaba2024qwen3,
  title={Qwen3 Technical Report: Scaling Multilingual Language Models},
  author={Alibaba Cloud},
  journal={Technical Report},
  year={2024}
}

@article{bai2025qwen2,
  title={Qwen2. 5-vl technical report},
  author={Bai, Shuai and Chen, Keqin and Liu, Xuejing and Wang, Jialin and Ge, Wenbin and Song, Sibo and Dang, Kai and Wang, Peng and Wang, Shijie and Tang, Jun and others},
  journal={arXiv preprint arXiv:2502.13923},
  year={2025}
}

@inproceedings{li2024multimodal,
  title={Multimodal ArXiv: A Dataset for Improving Scientific Comprehension of Large Vision-Language Models},
  author={Li, Lei and Wang, Yuqi and Xu, Runxin and Wang, Peiyi and Feng, Xiachong and Kong, Lingpeng and Liu, Qi},
  booktitle={Proceedings of the 62nd Annual Meeting of the Association for Computational Linguistics (Volume 1: Long Papers)},
  pages={14369--14387},
  year={2024}
}

@article{liang2025surveyx,
  title={Surveyx: Academic survey automation via large language models},
  author={Liang, Xun and Yang, Jiawei and Wang, Yezhaohui and Tang, Chen and Zheng, Zifan and Song, Shichao and Lin, Zehao and Yang, Yebin and Niu, Simin and Wang, Hanyu and others},
  journal={arXiv preprint arXiv:2502.14776},
  year={2025}
}

@article{wellawatte2025chemlit,
  title={ChemLit-QA: a human evaluated dataset for chemistry RAG tasks},
  author={Wellawatte, Geemi P and Guo, Huixuan and Lederbauer, Magdalena and Borisova, Anna and Hart, Matthew and Brucka, Marta and Schwaller, Philippe},
  journal={Machine Learning: Science and Technology},
  volume={6},
  number={2},
  pages={020601},
  year={2025},
  publisher={IOP Publishing}
}

@inproceedings{duan2025docopilot,
  title={Docopilot: Improving Multimodal Models for Document-Level Understanding},
  author={Duan, Yuchen and Chen, Zhe and Hu, Yusong and Wang, Weiyun and Ye, Shenglong and Shi, Botian and Lu, Lewei and Hou, Qibin and Lu, Tong and Li, Hongsheng and others},
  booktitle={Proceedings of the Computer Vision and Pattern Recognition Conference},
  pages={4026--4037},
  year={2025}
}

@inproceedings{
  zhou2025scientists,
  title={Scientists' First Exam: Probing Cognitive Abilities of {MLLM} via Perception, Understanding, and Reasoning},
  author={Yuhao Zhou and Yiheng Wang and Xuming He and Ruoyao Xiao and Zhiwei Li and Qiantai Feng and Zijie Guo and Yuejin Yang and Hao Wu and Wenxuan Huang and Jiaqi Wei and Dan Si and YAO XIUQI and Jia Bu and Haiwen Huang and Tianfan Fu and SHIXIANG TANG and Ben Fei and Dongzhan Zhou and Fenghua Ling and Yan Lu and Siqi Sun and Chenhui Li and Guanjie Zheng and Jiancheng Lv and Wenlong Zhang and LEI BAI},
  booktitle={The Thirty-ninth Annual Conference on Neural Information Processing Systems Datasets and Benchmarks Track},
  year={2025},
  url={https://openreview.net/forum?id=MhUendsT7L}
}

@inproceedings{hendrycks2021mmlu,
  title={Measuring Massive Multitask Language Understanding},
  author={Hendrycks, Dan and Burns, Collin and Basart, Steven and Zou, Andy and Mazeika, Mantas and Song, Dawn and Steinhardt, Jacob},
  booktitle={Proceedings of the International Conference on Learning Representations (ICLR)},
  year={2021}
}

@inproceedings{lu2022scienceqa,
  title={Learn to Explain: Multimodal Reasoning via Thought Chains for Science Question Answering},
  author={Lu, Pan and Mishra, Swaroop and Xia, Tony and Qiu, Liang and Chang, Kai-Wei and Zhu, Song-Chun and Tafjord, Oyvind and Clark, Peter and Kalyan, Ashwin},
  booktitle={Thirty-sixth Conference on Neural Information Processing Systems},
  year={2022}
}

@inproceedings{wang2023scibench,
  title={SciBench: Evaluating College-Level Scientific Problem-Solving Abilities of Large Language Models},
  author={Wang, Xiaoxuan and Hu, Ziniu and Lu, Pan and Zhu, Yanqiao and Zhang, Jieyu and Subramaniam, Satyen and Loomba, Arjun R. and Zhang, Shichang and Sun, Yizhou and Wang, Wei},
  booktitle={Proceedings of the 41st International Conference on Machine Learning (ICML)},
  year={2024}
}

@inproceedings{rein2023gpqa,
  title        = {{GPQA}: A Graduate-Level Google-Proof Q\&A Benchmark},
  author       = {Rein, David and Li Hou, Betty and Cooper Stickland, Asa and Petty, Jackson and Pang, Richard Yuanzhe and Dirani, Julien and Michael, Julian and Bowman, Samuel R.},
  booktitle    = {Proceedings of the 1st Conference on Language Modeling},
  year         = {2024}
}

@inproceedings{lu2023scieval,
  title={SciEval: A Multi-Level Large Language Model Evaluation Benchmark for Scientific Research},
  author={Sun, Liangtai and Han, Yang and Zhao, Zihan and Ma, Da and Shen, Zhennan and Chen, Baocai and Chen, Lu and Yu, Kai},
  booktitle={Proceedings of the AAAI Conference on Artificial Intelligence},
  year={2024}
}

@inproceedings{yue2024mmmu,
  title={MMMU: A Massive Multi-discipline Multimodal Understanding and Reasoning Benchmark for Expert AGI},
  author={Yue, Xiang and Ni, Yuansheng and Zhang, Kai and Zheng, Tianyu and Liu, Ruoqi and Zhang, Ge and Stevens, Samuel and Jiang, Dongfu and Ren, Weiming and Sun, Yuxuan and Wei, Cong and Yu, Botao and Yuan, Ruibin and Sun, Renliang and Yin, Ming and Zheng, Boyuan and Yang, Zhenzhu and Liu, Yibo and Huang, Wenhao and Sun, Huan and Su, Yu and Chen, Wenhu},
  booktitle={Proceedings of the IEEE/CVF Conference on Computer Vision and Pattern Recognition},
  pages={9556--9567},
  year={2024}
}

@article{wang2024autosurvey,
  title={Autosurvey: Large language models can automatically write surveys},
  author={Wang, Yidong and Guo, Qi and Yao, Wenjin and Zhang, Hongbo and Zhang, Xin and Wu, Zhen and Zhang, Meishan and Dai, Xinyu and Wen, Qingsong and Ye, Wei and others},
  journal={Advances in neural information processing systems},
  volume={37},
  pages={115119--115145},
  year={2024}
}

@article{feng2024sciknoweval,
  title={SciKnowEval: Evaluating Multi-level Scientific Knowledge of Large Language Models},
  author={Feng, Kehua and Shen, Xinyi and Wang, Weijie and Zhuang, Xiang and Tang, Yuqi and Zhang, Qiang and Ding, Keyan},
  journal={arXiv preprint arXiv:2406.09098},
  year={2024}
}

@article{map2025supergpqa,
  title={SuperGPQA: Scaling LLM Evaluation across 285 Graduate Disciplines},
  author={{M-A-P Team} and Du, Xinrun and Yao, Yifan and Ma, Kaijing and Wang, Bingli and Zheng, Tianyu and Zhu, King and Liu, Minghao and Liang, Yiming and Jin, Xiaolong and others},
  journal={arXiv preprint arXiv:2502.14739},
  year={2025}
}

@inproceedings{lai2023ds1000,
  title={DS-1000: A Natural and Reliable Benchmark for Data Science Code Generation},
  author={Lai, Yuhang and Li, Chengxi and Wang, Yiming and Zhang, Tianyi and Zhong, Ruiqi and Zettlemoyer, Luke and Yih, Wen-tau and Fried, Daniel and Wang, Sida and Yu, Tao},
  booktitle={Proceedings of the 40th International Conference on Machine Learning (ICML)},
  year={2023}
}

@inproceedings{liu2023agentbench,
  title={AgentBench: Evaluating LLMs as Agents},
  author={Liu, Xiao and Yu, Hao and Zhang, Hanchen and Xu, Yifan and Lei, Xuanyu and Lai, Hanyu and Gu, Yu and Ding, Hangliang and Men, Kaiwen and Yang, Kejuan and others},
  booktitle={Proceedings of the International Conference on Learning Representations (ICLR)},
  year={2024}
}

@inproceedings{jimenez2024swebench,
  title={SWE-bench: Can Language Models Resolve Real-World GitHub Issues?},
  author={Jimenez, Carlos E. and Yang, John and Wettig, Alexander and Yao, Shunyu and Pei, Kexin and Press, Ofir and Narasimhan, Karthik},
  booktitle={Proceedings of the International Conference on Learning Representations (ICLR)},
  year={2024}
}

@inproceedings{chen2025scienceagent,
  title={ScienceAgentBench: Toward Rigorous Assessment of Language Agents for Data-Driven Scientific Discovery},
  author={Chen, Ziru and Chen, Shijie and Ning, Yuting and Zhang, Qianheng and Wang, Boshi and Yu, Botao and Li, Yifei and Liao, Zeyi and Wei, Chen and Lu, Zitong and others},
  booktitle={Proceedings of the International Conference on Learning Representations (ICLR)},
  year={2025}
}

@inproceedings{hendrycks2021measuring,
  title={Measuring Massive Multitask Language Understanding},
  author={Hendrycks, Dan and Burns, Collin and Basart, Steven and Zou, Andy and Mazeika, Mantas and Song, Dawn and Steinhardt, Jacob},
  booktitle={International Conference on Learning Representations},
  year={2021}
}

@inproceedings{wang2024scibench,
  title={SciBench: Evaluating College-Level Scientific Problem-Solving Abilities of Large Language Models},
  author={Wang, Xiaoxuan and Hu, Ziniu and Lu, Pan and Zhu, Yanqiao and Zhang, Jieyu and Subramaniam, Satyen and Loomba, Arjun R. and Zhang, Shichang and Sun, Yizhou and Wang, Wei},
  booktitle={Proceedings of the 41st International Conference on Machine Learning},
  pages={50622--50649},
  year={2024},
  volume={235},
  series={Proceedings of Machine Learning Research},
  publisher={PMLR}
}

@inproceedings{wang2024mmlupro,
  title={MMLU-Pro: A More Robust and Challenging Multi-Task Language Understanding Benchmark},
  author={Wang, Yubo and Ma, Xueguang and Zhang, Ge and Ni, Yuansheng and Chandra, Abhranil and Guo, Shiguang and Ren, Weiming and Arulraj, Aaran and He, Xuan and Jiang, Ziyan and Li, Tianle and Ku, Max and Wang, Kai and Zhuang, Alex and Fan, Rongqi and Xue, Xiang and Chen, Wenhu},
  booktitle={Thirty-eighth Conference on Neural Information Processing Systems Datasets and Benchmarks Track},
  year={2024}
}

@inproceedings{huber2025llms,
  title={LLMs meet Bloom's Taxonomy: A Cognitive View on Large Language Model Evaluations},
  author={Huber, Thomas and Niklaus, Christina},
  booktitle={Proceedings of the 31st International Conference on Computational Linguistics},
  pages={5211--5246},
  year={2025}
}

@inproceedings{tian2024scicode,
  title={SciCode: A Research Coding Benchmark Curated by Scientists},
  author={Tian, Minyang and Gao, Luyu and Zhang, Shizhuo Dylan and Chen, Xinan and others},
  booktitle={Thirty-eighth Conference on Neural Information Processing Systems Datasets and Benchmarks Track},
  year={2024}
}

@inproceedings{li2024m3sciqa,
  title={M3SciQA: A Multi-Modal Multi-Document Scientific QA Benchmark for Evaluating Foundation Models},
  author={Li, Chuhan and Shangguan, Ziyao and Zhao, Yilun and Li, Deyuan and Liu, Yixin and Cohan, Arman},
  booktitle={Findings of the Association for Computational Linguistics: EMNLP 2024},
  pages={15419--15446},
  year={2024}
}

@inproceedings{li2024multimodalarxiv,
  title={Multimodal ArXiv: A Dataset for Improving Scientific Comprehension of Large Vision-Language Models},
  author={Li, Lei and Wang, Yuqi and Xu, Runxin and Wang, Peiyi and Feng, Xiachong and Kong, Lingpeng and Liu, Qi},
  booktitle={Proceedings of the 62nd Annual Meeting of the Association for Computational Linguistics (Volume 1: Long Papers)},
  pages={14369--14387},
  year={2024}
}

@inproceedings{DBLP:conf/aaaifs/TsatsaronisSPAAGGAAZN12,
  author       = {George Tsatsaronis and
                  Michael Schroeder and
                  Georgios Paliouras and
                  Yannis Almirantis and
                  Ion Androutsopoulos and
                  {\'{E}}ric Gaussier and
                  Patrick Gallinari and
                  Thierry Arti{\`{e}}res and
                  Michael R. Alvers and
                  Matthias Zschunke and
                  Axel{-}Cyrille Ngonga Ngomo},
  title        = {BioASQ: {A} Challenge on Large-Scale Biomedical Semantic Indexing
                  and Question Answering},
  booktitle    = {Information Retrieval and Knowledge Discovery in Biomedical Text,
                  Papers from the 2012 {AAAI} Fall Symposium, Arlington, Virginia, USA,
                  November 2-4, 2012},
  series       = {{AAAI} Technical Report},
  volume       = {{FS-12-05}},
  publisher    = {{AAAI}},
  year         = {2012},
  url          = {http://www.aaai.org/ocs/index.php/FSS/FSS12/paper/view/5600},
  bibsource    = {dblp computer science bibliography, https://dblp.org}
}

@inproceedings{chen2024internvl,
  title={Internvl: Scaling up vision foundation models and aligning for generic visual-linguistic tasks},
  author={Chen, Zhe and Wu, Jiannan and Wang, Wenhai and Su, Weijie and Chen, Guo and Xing, Sen and Zhong, Muyan and Zhang, Qinglong and Zhu, Xizhou and Lu, Lewei and others},
  booktitle={Proceedings of the IEEE/CVF Conference on Computer Vision and Pattern Recognition},
  pages={24185--24198},
  year={2024}
}

@misc{vteam2025glm45vglm41vthinkingversatilemultimodal,
      title={GLM-4.5V and GLM-4.1V-Thinking: Towards Versatile Multimodal Reasoning with Scalable Reinforcement Learning}, 
      author={V Team and Wenyi Hong and Wenmeng Yu and Xiaotao Gu and Guo Wang and Guobing Gan and Haomiao Tang and Jiale Cheng and Ji Qi and Junhui Ji and Lihang Pan and Shuaiqi Duan and Weihan Wang and Yan Wang and Yean Cheng and Zehai He and Zhe Su and Zhen Yang and Ziyang Pan and Aohan Zeng and Baoxu Wang and Bin Chen and Boyan Shi and Changyu Pang and Chenhui Zhang and Da Yin and Fan Yang and Guoqing Chen and Jiazheng Xu and Jiale Zhu and Jiali Chen and Jing Chen and Jinhao Chen and Jinghao Lin and Jinjiang Wang and Junjie Chen and Leqi Lei and Letian Gong and Leyi Pan and Mingdao Liu and Mingde Xu and Mingzhi Zhang and Qinkai Zheng and Sheng Yang and Shi Zhong and Shiyu Huang and Shuyuan Zhao and Siyan Xue and Shangqin Tu and Shengbiao Meng and Tianshu Zhang and Tianwei Luo and Tianxiang Hao and Tianyu Tong and Wenkai Li and Wei Jia and Xiao Liu and Xiaohan Zhang and Xin Lyu and Xinyue Fan and Xuancheng Huang and Yanling Wang and Yadong Xue and Yanfeng Wang and Yanzi Wang and Yifan An and Yifan Du and Yiming Shi and Yiheng Huang and Yilin Niu and Yuan Wang and Yuanchang Yue and Yuchen Li and Yutao Zhang and Yuting Wang and Yu Wang and Yuxuan Zhang and Zhao Xue and Zhenyu Hou and Zhengxiao Du and Zihan Wang and Peng Zhang and Debing Liu and Bin Xu and Juanzi Li and Minlie Huang and Yuxiao Dong and Jie Tang},
      year={2025},
      eprint={2507.01006},
      archivePrefix={arXiv},
      primaryClass={cs.CV},
      url={https://arxiv.org/abs/2507.01006}, 
}

@article{tongyidr,
  title={Tongyi DeepResearch Technical Report},
  author={Team, Tongyi DeepResearch and Li, Baixuan and Zhang, Bo and Zhang, Dingchu and Huang, Fei and Li, Guangyu and Chen, Guoxin and Yin, Huifeng and Wu, Jialong and Zhou, Jingren and others},
  journal={arXiv preprint arXiv:2510.24701},
  year={2025}
}

\section*{Author Contributions}

The HiSciBench benchmark was collaboratively developed, with author responsibilities organized by 5 level and corresponding scientific tasks.

\noindent\textbf{L1 Scientific Literacy Evaluation:} 
{Xingquan Zhang} led the design, annotation, and validation of 1,200 factual and reasoning questions across six scientific disciplines.

\noindent\textbf{L2 Document Understanding and Translation:} 
{Yupu Liang} and {Zhiyuan Chen} were responsible for 
{L2.1: Document Parsing and OCR} 
and {L2.2: Cross-lingual Translation},respectively.

\noindent\textbf{L3 Literature Question Answering:} 
{Qixuan Zhang} developed {L3.1: Monolingual Literature QA}  and {Zhiyuan Chen} developed
{L3.2: Cross-lingual Literature QA} , 
including corpus construction and evaluation design.

\noindent\textbf{L4 Topic-guided Literature Review:}        
{Wenwen Zhuang} led the construction of 60 review topics across six disciplines and the development of evaluation protocols for multi-document synthesis.

\noindent\textbf{L5 Data-driven Scientific Discovery:} 
{Xingquan Zhang} directed the task design and data curation for 74 computational discovery tasks across four disciplines.

\noindent\textbf{Overall Coordination and Writing:} 
{Yaping Zhang}, {Yang Zhao}, and {Lu Xiang} coordinated the project execution, organized the manuscript structure, and contributed to the 
{Abstract}, {Introduction}, and {Conclusion} sections.

\noindent\textbf{Supervision and Review:} 
Jiajun Zhang, Yu Zhou, and Chengqing Zong provided overall supervision, theoretical guidance, and critical review, ensuring the methodological rigor and consistency of the study.

All authors discussed the results, contributed to manuscript refinement, and approved the final submission.

\end{document}